\documentclass{article}

\usepackage[nonatbib, preprint]{neurips_2022}

\usepackage[square, numbers]{natbib}
\usepackage[utf8]{inputenc} 
\usepackage[T1]{fontenc}    
\usepackage{hyperref}       
\usepackage{url}            
\usepackage{booktabs}       
\usepackage{amsfonts}       
\usepackage{nicefrac}       
\usepackage{microtype}      
\usepackage{multirow}

\usepackage{amsmath,amsfonts,bm}

\def\eqref#1{equation~\ref{#1}}

\def\1{\bm{1}}

\DeclareMathAlphabet{\mathsfit}{\encodingdefault}{\sfdefault}{m}{sl}
\SetMathAlphabet{\mathsfit}{bold}{\encodingdefault}{\sfdefault}{bx}{n}

\def\gG{{\mathcal{G}}}

\def\sN{{\mathbb{N}}}

\def\sS{{\mathbb{S}}}

\def\sV{{\mathbb{V}}}

\newcommand{\softmax}{\mathrm{softmax}}

\DeclareMathOperator*{\argmax}{arg\,max}
\DeclareMathOperator*{\argmin}{arg\,min}

\usepackage{amssymb} 

\usepackage[ruled,vlined]{algorithm2e}
\SetKwInput{KwInput}{Input} 
\SetKwInput{KwOutput}{Output} 

\usepackage{enumitem}

\usepackage{bbm}

\usepackage{mathtools}

\usepackage[usenames, dvipsnames]{xcolor}
\hypersetup{colorlinks,citecolor=NavyBlue,
linkcolor=NavyBlue,urlcolor=NavyBlue}
\usepackage[nameinlink,capitalise]{cleveref}

\usepackage{tikz}

\usepackage[toc,page,header]{appendix}
\usepackage{minitoc}

\newcommand{\model}[1]{{\fontfamily{lmtt}\selectfont{#1}}}

\title{Neuro CROSS exchange: Learning to CROSS exchange to solve realistic vehicle routing problems}

\author{
    Minjun Kim$^\dagger$, Junyoung Park$^\dagger$, Jinkyoo Park$^*$\\
    KAIST \\
    \texttt{$\{$minjun1212, junyoungpark, jinkyoo.park$\}$@kaist.ac.kr}
}

\begin{document}

\maketitle

\def\thefootnote{$\dagger$}\footnotetext{Equal contribution}\def\thefootnote{\arabic{footnote}}

\def\thefootnote{*}\footnotetext{Corresponding author}\def\thefootnote{\arabic{footnote}}

\doparttoc
\faketableofcontents

\begin{abstract}
CROSS exchange (CE), a meta-heuristic that solves various vehicle routing problems (VRPs), improves the solutions of VRPs by swapping the sub-tours of the vehicles. Inspired by CE, we propose Neuro CE (NCE), a fundamental operator of \textit{learned} meta-heuristic, to solve various VRPs while overcoming the limitations of CE (i.e., the expensive $\mathcal{O}(n^4)$ search cost). NCE employs graph neural network to predict the cost-decrements (i.e., results of CE searches) and utilizes the predicted cost-decrements as guidance for search to decrease the search cost to $\mathcal{O}(n^2)$. As the learning objective of NCE is to predict the cost-decrement, the training can be simply done in a supervised fashion, whose training samples can be prepared effortlessly. Despite the simplicity of NCE, numerical results show that the NCE trained with flexible multi-depot VRP (FMDVRP) outperforms the meta-heuristic baselines. More importantly, it significantly outperforms the neural baselines when solving distinctive special cases of FMDVRP (e.g., MDVRP, mTSP, CVRP) without additional training.
\end{abstract}

\section{Introduction}
The field of neural combinatorial optimization (NCO), an emerging research area intersecting operation research and artificial intelligence, aims to train an effective solver for various combinatorial optimization, such as the traveling salesman problem (TSP) \citep{bello2016neural, khalil2017learning, nazari2018reinforcement, kool2018attention,kwon2020pomo}, vehicle routing problems (VRPs) \citep{bello2016neural, khalil2017learning, nazari2018reinforcement, kool2018attention,kwon2020pomo,hottung2019neural, lu2019learning, da2021learning}, and vertex covering problems \citep{khalil2017learning,li2018combinatorial,guo2019solving}. As NCO tackles NP-hard problems using various state-of-the-art (SOTA) deep learning techniques, it is considered an important research area in artificial intelligence. At the same time, NCO is an important field from a practical point of view because it can solve complex real-world problems.

Most NCO methods learn an operator that improves the current solution to obtain a better solution (i.e., improvement heuristics) \citep{hottung2019neural, lu2019learning, da2021learning} or constructs a solution sequentially (i.e., construction heuristics) \citep{bello2016neural, khalil2017learning, nazari2018reinforcement,kool2018attention,kwon2020pomo,park2021schedulenet,cao2021dan}. To learn such operators, NCO methods either employ supervised learning (SL) (which imitates the solutions of the verified solvers) or reinforcement learning (RL) (which necessitates the design of an effective representation, architecture or learning method), making them less trainable for complex and realistic VRPs. Moreover, most NCO researches in recent years has extensively focused on improving the performance of the benchmark CO problems while overlooking the applicability of NCO to more realistic problems. 




Focusing on that \textit{improvement} (meta) heuristics are applicable various VRP with some minor problem-specific modifications, we aim to learn a fundamental and universal improvement operator that overcomes the limitation of CROSS-exchange (CE) \cite{taillard1997tabu}, a generalization of various hand-craft improvement operators of meta heuristics. CE improves the solution of VRP by updating the tours of two vehicles. To be specific, it chooses the sub-tours from each tour and swap the sub-tours to generate the updated tours. In practice, to find the (best) \textit{improving} sub-tours (i.e., the sub-tours that decrease the cost value of VRP), CE performs brute-force search that costs $\mathcal{O}(N^4)$, which makes CE unsuitable for large scale VRPs.  




In this paper, we propose Neuro CE (NCE) that effectively conducts the CE operation with significantly less computational complexity. NCE amortizes the search for ending nodes of the sub-tours by employing a 
graph neural network (GNN) that predicts the best cost decrement, given two starting nodes from the given two trajectories. By using the predictions, NCE searches over the promising starting nodes only. Hence, the proposed NCE has $\mathcal{O}(N^2)$ search complexity. Furthermore, unlike other SL or RL approaches, the prediction target of NCE is not the entire solution of VRP, but the cost decrements of the CE operations that lowers the difficulty of the prediction task. This allows the training data to be prepared effortlessly.

The contributions of this study are summarized as follows:
\begin{itemize}[leftmargin=0.5cm]
\vspace{-0.25cm}
    \item \textbf{Generalizability/Transferability:} As NCE learns a fundamental and universal operator, it can solve various complex VRPs without training for each type of VRPs without retraining.
    \item \textbf{Trainability:} The NCE operator is trained in a supervised manner with the dataset comprised of the tour pairs and cost decrements, which are easy to obtain.
    \item \textbf{Practicality/Performance:} 
    We evaluate NCE with various types of VRPs, including flexible multi-depot VRP (FMDVRP), multi-depot VRP (MDVRP), multiple traveling salesman problem (mTSP), and capacitated VRP (CVRP). Extensive numerical experiments validate that the strong empirical performance of NCE compared to the SOTA meta-heuristics and NCO baselines even though NCE is only trained to solve FMDVRP.
\end{itemize}

\section{Preliminaries}

\begin{figure}[t]
    \centering
    \includegraphics[width=\linewidth]{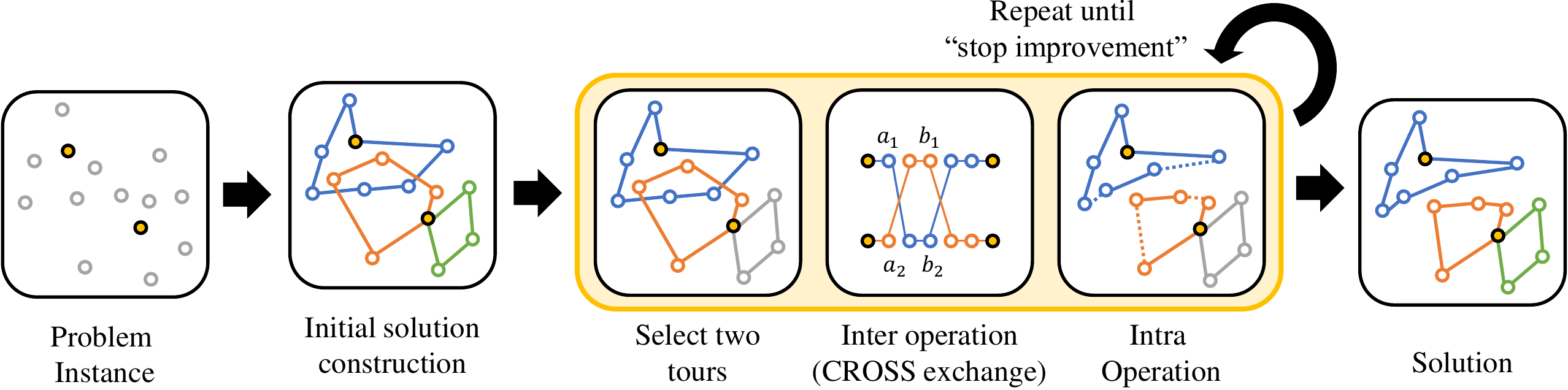}
    \caption{The overall procedure of improvement heuristic that uses CE as the inter-operation.}
    \label{fig:ce_overall}
\end{figure}

This section introduces the target problem, flexible multi-depot VRP (FMDVRP) and CE, which is one of the possible approach that solves FMDVRP.

\subsection{Min-max flexible multi-depot VRP}
\label{subsec:fmdvrp}
Min-max FMDVRP is a generalization of VRP that aims to find the coordinated routes of multiple vehicles with multiple depots. The flexibility allows vehicles to go back to any depots regardless of their starting depots. FMDVRP is formulated as follows:
\begin{align}
\label{eqn:vrp_obj}
\min_{\pi \in \sS(P)}\max_{i \in \sV}{C(\tau_i)}
\end{align}
where $P$ is the description of the FMDVRP instance that is composed of a set of vehicles $\sV$, $\sS(P)$ is the set of solutions that satisfy the constraints of FMDVRP (i.e., feasible solutions), and  $\pi=\{\tau_i\}_{i \in \sV}$ is a solution of the min-max FMDVRP. The tour  $\tau_i=[N_1, N_2, ..., N_{l(i)}]$ of vehicle $i$ is the ordered collection of the visited cities by the vehicle $v_i$, $C(\tau_i)$ is the cost of $\tau_i$. FMDVRP reflects VRP where the vehicles are shared and pickup/delivered from arbitrary space (e.g., shared rental car services). For the mixed integer linear programming (MILP) formulation of FMDVRP, please refer to \cref{appendix:fmdvrp}.

Classical VRPs are special cases of FMVDRP. \emph{TSP} is a VRP with a single vehicle and depot, \emph{mTSP} is a VRP with multiple vehicles and a single depot, and \emph{MDVRP} is a VRP with multiple vehicles and depots. Since FMVDRP is a general problem class, we learn a solver for FMVDRP and employ it to solve other specific problems (i.e., MDVRP, mTSP, and CVRP), without retraining or fine-tuning. We demonstrate that the proposed method can solve the special cases without retraining in \cref{sec:experiments}.

\begin{algorithm}[t!]
\caption{Neuro CROSS exchange (NCE) for solving VRP family}
\label{alg:NCE-overall}
\DontPrintSemicolon
  \KwInput{VRP instance $P$, cost-decrement prediction model $f_\theta$, Perturbation parameter $p$}
  \KwOutput{Optimized tours $\{\tau_i^*\}_{i\in|\sV|}$}
  
$\{\tau_i\}_{i\in|\sV|} \leftarrow \text{\fontfamily{lmtt}\selectfont GetInitialSolution}(P)$ \\
$C_\text{per} \leftarrow 0$ \\

\While{True}{
  \While{improvement}
  { 
    $(\tau_1, \tau_2) \leftarrow \text{\fontfamily{lmtt}\selectfont SelectTours}(\{\tau_i\}_{i\in|\sV|})$ \\
    $(\tau_1', \tau_2') \leftarrow \text{\fontfamily{lmtt}\selectfont NeuroCROSS}(\tau_1, \tau_2, f_\theta)$ \tcp*{Inter operation}
    $\tau_i' \leftarrow \text{\fontfamily{lmtt}\selectfont IntraOperation}(\tau_i), i=1,2$\\
    $\tau_1 \leftarrow \tau_1', \tau_2 \leftarrow \tau_2'$
  }
  \If{update}
  {$\{\tau^*_i\}_{i\in|\sV|} \leftarrow \{\tau_i\}_{i\in|\sV|}$}
  \If{$C_\text{per}=p$}
  {\textbf{break}}
  $C_\text{per} \leftarrow C_\text{per}+1$ \\
  $(\tau_1, \tau_2) \leftarrow \text{ChooseRandomTours}$ \\
  $(\tau_1, \tau_2) \leftarrow \text{RandomExchange}(\tau_1, \tau_2)$ \tcp*{escape from local minima}
  }

\end{algorithm}


\subsection{CROSS exchange}
CE is a solution updating operator that iteratively improves the solution until it reaches a satisfactory result \cite{taillard1997tabu}. CE reduces the overall cost by \textit{exchanging} the sub-tours in two tours. The CE operator is defined as:
\begin{align}
\label{eqn:cross-exchange}
\tau_1', \tau_2' = \text{\fontfamily{lmtt}\selectfont CROSS}(a_1, b_1, a_2, b_2;\tau_1,\tau_2) \\
\tau_1' \triangleq \tau_1[:a_1] \oplus \tau_2[a_2:b_2] \oplus \tau_1[b_1:] \\
\tau_2' \triangleq \tau_2[:a_2] \oplus \tau_1[a_1:b_1] \oplus \tau_2[b_2:]
\end{align}
where $\tau_i$ and $\tau_i'$ are the input and updated tours of the vehicle $i$, respectively. $\tau_i[a:b]$ represents the sub-tour of $\tau_i$, ranging from node $a$ to $b$. $\tau \oplus \tau'$ represents the concatenation of tours $\tau$ and $\tau'$. For brevity, we assume the node $a_1, a_2$ comes early than node $b_1, b_2$ in $\tau_1, \tau_2$ , respectively.

CE selects the sub-tours (i.e., $\tau_1[a_1:b_1], \tau_2[a_2:b_2]$) from $\tau_1,\tau_2$ and swaps the sub-tours to generate new tours $\tau_1', \tau_2'$. CE seeks to find the four points $(a_1, b_1, a_2, b_2)$ to reduce the cost of the tours (i.e., $\max(C(\tau_1'), C(\tau_2')) \leq \max(C(\tau_1), C(\tau_2))$). When the full search method is naively employed, the search cost is $\mathcal{O}(n^4)$, where $n$ is the number of nodes in a tour.


\cref{fig:ce_overall} illustrates how improvement heuristics utilize CE to solve FMDVRP. The improvement heuristics start by generating the initial feasible tours using simple heuristics. Then, they repeatedly (1) select two tours, (2) apply inter-operation to generate improved tours by CE, and (3) apply intra-operation to improve the tours independently. The application of the inter-operation makes the heuristics more suitable for solving the multi-vehicle routing problems as it considers the interactions among the vehicles while \textit{improving} the solutions. The improvement heuristics terminate when no more (local) improvement is possible.

\section{Neuro CROSS Exchange}
\label{sec:nce}

In this section, we introduce Neuro CROSS exchange (NCE) to solve FMDVRP and its special cases. The overall procedure of NCE is summarized in \cref{alg:NCE-overall}. We briefly explain
\model{GetInitialSolution}, \model{SelectTours}, \model{NeuroCROSS}, and \model{IntraOperation}, and then provide the details of the proposed {\fontfamily{lmtt}\selectfont NeuroCROSS} operation in the following subsections. NCE is particularly designed to enhance CE to improve the solution quality and solving speed. Each component of NCE is as follows:
\begin{itemize}[leftmargin=0.5cm]
\vspace{-0.25cm}
    \item \textbf{\model{GetInitialSolution} } We use a multi-agent extended version of the greedy assignment heuristic to obtain the initial feasible solutions. The heuristic first clusters the cities into $|\sV|$ clusters and then applies the greedy assignment to each cluster to get the initial solution. 
    \item \textbf{\model{SelectTours} } Following the common practice, we set $\tau_1, \tau_2$ as the tours of the largest and smallest cost (i.e., $\tau_1 = \argmax_{\tau}({C(\tau_i)}_{i\in\sV})$, $\tau_2 = \argmin_{\tau}({C(\tau_i)}_{i\in\sV})$).
    \item \textbf{\model{NeruoCROSS} } We utilize the cost-decrement prediction model $f_\theta(\cdot)$ and two-stage search method to find the cost-improving tour pair $(\tau_1',\tau_2')$ with $2\mathcal{O}(n^2)$ budget. The details of NCE operation will be given in \cref{subsec:NCE,subsec:cost-decr-pred}.
    \item \textbf{\model{IntraOperation} } For our targeting VRPs, the intra operation is equivalent to solving traveling salesman problem (TSP). We utilize {\fontfamily{lmtt}\selectfont elkai} \cite{elkai} to solve TSP.
\end{itemize}
\vspace{-0.25cm}

\subsection{Neuro CROSS exchange operation}
\label{subsec:NCE}
\begin{algorithm}[t]
\caption{{\fontfamily{lmtt}\selectfont NeuroCROSS} operation}
\label{alg:NCE-op}
\DontPrintSemicolon
  \KwInput{tours $\tau_1, \tau_2$, cost-decrement prediction model $f_\theta(\cdot)$}
  \KwOutput{updated tours $\tau'_1, \tau'_2$}
  \tcc{Predict cost decrement}
  $\sS \leftarrow \{\emptyset\}$ \\
  \For{$(a_1, a_2) \in {\tau_1} \times {\tau_2}$}{
    $\hat y^*(a_1,a_2; \tau_1, \tau_2) \leftarrow f_\theta(a_1, a_2; \tau_1, \tau_2)$ \tcp*{Cost-decrement prediction}
    $\sS \leftarrow \sS \cup \{\left((a_1,a_2), \hat y^*(a_1,a_2; \tau_1, \tau_2)\right)\}$
    }
    
  \tcc{Candidate set construction}
  Sort $\sS$ by $y^*(a_1,a_2; \tau_1, \tau_2)$ in the descending order \\
  $\sS_K \leftarrow$ Take first $K$ elements of $\sS$ \\ 
  
  \tcc{Perform search}
  $a^*_1 \leftarrow \emptyset, a^*_2 \leftarrow \emptyset, 
  b^*_1 \leftarrow \emptyset, b^*_2 \leftarrow \emptyset,
  y^* \leftarrow 0$ \\
  \For{$\left((a_1,a_2), \hat y^*(a_1,a_2; \tau_1, \tau_2)\right) \in \sS_K$}{
    $(\bar b_1, \bar b_2)\leftarrow\argmax_{b_1, b_2}\left(C(\text{\fontfamily{lmtt}\selectfont CROSS}((a_1,b_1,a_2,b_2; \tau_1, \tau_2)))-C(\tau_1, \tau_2)\right)$ \\
    $y^*(a_1,a_2;\tau_1, \tau_2)\leftarrow C(\text{\fontfamily{lmtt}\selectfont CROSS}((a_1, \bar b_1,a_2,\bar b_2; \tau_1, \tau_2))-C(\tau_1, \tau_2)$ \\
  \If{$y^*(a_1,a_2; \tau_1, \tau_2) \geq y^*$}
        {$a^*_1 \leftarrow a_1, a^*_2 \leftarrow a_2$,
         $b^*_1 \leftarrow \bar b_1, b^*_2 \leftarrow \bar b_2$ \\
        $y^* \leftarrow y^*(a_1,a_2; \tau_1, \tau_2)$}
  }
  $(\tau_1',\tau'_2) \leftarrow \text{\fontfamily{lmtt}\selectfont CROSS}(a^*_1,b^*_1,a^*_2,b^*_2; \tau_1, \tau_2)$

\end{algorithm}
\vskip -0.1in

The CE operation can be shown as selecting two pairs of nodes (i.e., the pairs of $a_1/b_1$ and $a_2/b_2$) from the selected tours (i.e., $\tau_1, \tau_2$). This typically involves $\mathcal{O}(n^4)$ searches. To reduce the high search complexity, NCE utilizes the cost-decrement model $f_\theta(a_1, a_2; \tau_1, \tau_2)$ that predicts the maximum cost decrements from the given $\tau_1$ and $\tau_2$, and the starting nodes $a_1$ and $a_2$ of their sub-tours. That is, $f_\theta(a_1, a_2; \tau_1, \tau_2)$ amortizes the search for the ending nodes $b_1, b_2$ given $(\tau_1, \tau_2, a_1, a_2)$, and it helps to identify the promising $(a_1,a_2)$ pairs that are likely improve tours. After selecting the top $K$ promising pairs of $(a_1,a_2)$ using $f_\theta(a_1, a_2; \tau_1, \tau_2)$, whose search cost is $\mathcal{O}(n^2)$, NCE then finds ($b_1, b_2$) to identify the promising $(a_1,a_2)$ pairs. Overall, the entire search can be done in $2\mathcal{O}(n^2)$. The following paragraphs detail the procedures of NCE.

\paragraph{Predicting cost decrement} We employ $f_\theta(a_1, a_2; \tau_1, \tau_2)$ (which will be explained in Section \ref{subsec:cost-decr-pred}) to predict the optimal cost decrement $y^*$ defined as: 
\begin{align}
\label{eqn:NCE-opt-dec-cost}
y^*(a_1, a_2; \tau_1, \tau_2) &\coloneqq \max_{b_1, b_2}\left(C(\text{\fontfamily{lmtt}\selectfont CROSS}((a_1,b_1,a_2,b_2; \tau_1, \tau_2)))-C(\tau_1, \tau_2)\right)  \\
&\approx f_\theta(a_1, a_2; \tau_1, \tau_2)
\end{align}
where $C(\tau_1, \tau_2)$ is a shorthand notation of $\max\left(C(\tau_1), C(\tau_2)\right)$. In other words,  $f_\theta(\cdot)$ predicts the best cost decrement of $\tau_1$ and $\tau_2$, given $a_1$ and $a_2$ (i.e., the results of search algorithm), respectively.

\paragraph{Constructing search candidate set } By training $f_\theta(\cdot)$, we can amortize the search for $b_1$ and $b_2$. However, this amortization bears the prediction errors, which can misguide entire improvement process. To alleviate this problem, we select the top $K$ pairs of $(a_1, a_2)$ that have the largest $y^*$ out of all $(a_1,a_2)$ choices. Intuitively speaking, NCE exclude the less promising $(a_1,a_2)$ pairs while considering the prediction error of $f_\theta(\cdot)$ by allowing the following search for the top $K$ pairs.

\paragraph{Performing reduced search } NCE finds the best $(b_1, b_2)$ for each $(a_1, a_2)$ in the search candidate set and select the best cost decreasing $(a_1, a_2, b_1, b_2)$. Unlike the full search of CE, the proposed NCE only performs the search for $(b_1, b_2)$. This reduces the search cost from $\mathcal{O}(n^4)$ to $\mathcal{O}(n^2)$. The detailed procedures of NCE are summarized in \cref{alg:NCE-op}.

\subsection{Cost-decrement prediction model}
\label{subsec:cost-decr-pred}

\begin{figure}[t]
    \centering
    \includegraphics[width={\linewidth}]{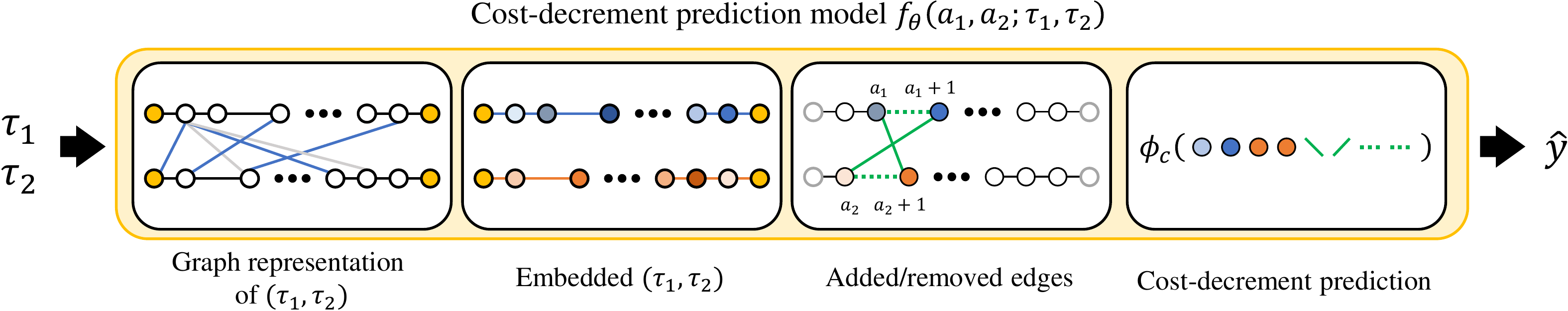}
    \caption{Cost-decrement prediction procedure}
    \label{fig:cost-decr-pred}
\end{figure}

NCE saves computations by employing $f_\theta(a_1,a_2;\tau_1,\tau_2)$ to predict $y^*(\cdot)$ from $a_1,a_2,\tau_1$ and $\tau_2$. The overall procedure is illustrated in \cref{fig:cost-decr-pred}.

\paragraph{Graph representation of $(\tau_1, \tau_2)$}
We represent the pair of tours $(\tau_1, \tau_2)$ as the directed complete graph $\gG=(\sN, \mathbb{E})$, where $\sN=\tau_1 \cup \tau_2$ (i.e., the $i^{\text{th}}$ node $n_i$ of $\gG$ is either the city or depot of the tours, and $e_{ij}$ is the edge from $n_i$ to $n_j$). $\gG$ has the following node and edge features:
\begin{itemize}[leftmargin=0.5cm]
\vspace{-0.25cm}
    \item $x_i \triangleq [\textbf{coord}(n_i), \mathbbm{1}_{\textbf{depot}}(n_i)]$, where $\textbf{coord}(n_i)$ is the 2D Euclidean coordinate of $v_i$, and $\mathbbm{1}_{\textbf{depot}}(n_i)$ is the indicator of whether $n_i$ is a depot.
    \item $x_{ij} \triangleq [\textbf{dist}(n_i, n_j)]$, where $\textbf{dist}(n_i, n_j)$ is the 2D Euclidean distance between $n_i$ and $n_j$.
\end{itemize}
\vspace{-0.25cm}

\paragraph{Graph embedding with attentive graph neural network (GNN)} We employ an attentive variant of graph-network (GN) block \cite{battaglia2018relational} to embed $\gG$. The attentive embedding layer is defined as follows:
\begin{align}
\label{eq:gnn-layer}
h'_{ij} &= \phi_e(h_i,h_j,h_{ij},x_{ij}) \\
z_{ij} &= \phi_w(h_i,h_j,h_{ij},x_{ij}) \\ 
w_{ij} &= \softmax(\{z_{ij}\}_{j \in \mathcal{N}(i)}) \\    
h'_{i} &= \phi_n(h_i, \sum_{j \in \mathcal{N}(i)} w_{ij} h'_{ij})
\end{align}
where $h_i$ and $h_{ij}$ are node and edge embeddings respectively, $\phi_e$, $\phi_w$, and $\phi_n$ are the Multilayer Perceptron (MLP)-parameterized edge, attention and node operators respectively, and $\mathcal{N}(i)$ is the neighbor set of $n_i$. We utilize $H$ embedding layers to compute the final node  $\{h^{(H)}_i |\, n_i\in \sV\}$ and edge embeddings $\{h^{(H)}_{ij} | \, e_{ij}\in \mathbb{E}\}$.

\textbf{Cost-decrement prediction } Based on the computed embedding, the cost prediction module $\phi_c$ predicts $y^*(a_1,a_2;\tau_1,\tau_2)$. The selection of the two starting nodes in $\tau_1$ and $\tau_2$ indicates (1) the addition of the two edges, $(a_1, a_2+1)$ and $(a_2, a_1+1)$, and (2) the removal of the original two edges, $(a_1, a_1+1)$ and $(a_2, a_2+1)$, as shown in the third block in \cref{fig:cost-decr-pred} (We overload the notation $a_1+1, a_2+1$ so that they denote the next nodes of $a_1, a_2$ in $\tau_1, \tau_2$, respectively). To consider such edge addition and removal procedure in cost prediction, we design $\phi_c$ as follows:
\begin{align}
\label{eq:cost-pred-layer}
\hat y^*(a_1,a_2;\tau_1,\tau_2)=\phi_c(
\underbrace{h^{(H)}_{a_1}, 
            h^{(H)}_{a_1+1}, 
            h^{(H)}_{a_2}, 
            h^{(H)}_{a_2+1}}_{\tikz\draw[NavyBlue, fill=NavyBlue] (0,0) circle (.5ex); /
                              \tikz\draw[Orange, fill=Orange] (0,0) circle (.5ex);
                              \textbf{ : node embedding}},
\underbrace{h^{(H)}_{a_1,a_2+1}, h^{(H)}_{a_2,a_1+1}}_
      {\tikz\draw[ForestGreen, thick] (0,1.0)--(0.3,1.0); 
       \textbf{ : link addition}},
\underbrace{h^{(H)}_{a_1,a_1+1}, h^{(H)}_{a_2,a_2+1}}_
      {\tikz\draw[ForestGreen, thick, densely dotted] (0,1.0)--(0.3,1.0);
      \textbf{ : link removal}}
)
\end{align}
where $h^{(H)}_i$ and $h^{(H)}_{i,j}$ denotes the embedding of $n_i$ and $e_{ij}$, respectively.

The quality of NCE operator highly depends on the accuracy of $f_\theta$. When $K \geq {10}$, we experimentally confirmed that the NCE operator finds the argmax $(a_1,a_2,b_1,b_2)$ pair with high probability. We provide the experimental details and results about the predictions of $f_\theta$ in \cref{appendix:train}.

\section{Related works}
\paragraph{Supervised learning (SL) approach to solve VRPs} SL approaches \cite{joshi2019efficient, vinyals2015pointer, xin2021neurolkh,li2021learning,li2018combinatorial} utilize the supervision from the VRP solvers as the training labels. \cite{vinyals2015pointer, joshi2019efficient} imitates TSP solvers using PointerNet and graph convolution network (GCN), respectively. \cite{joshi2019efficient} trains a GCN to predict the edge occurrence probabilities in TSP solutions. Even though SL often offer a faster solving speed than existing solvers, their use is limited to the problems where the solvers are available. Such property limits the use of SL from general and realistic VRPs.

\paragraph{Reinforcement learning (RL) approach to solve VRPs} 
RL approaches \citep{bello2016neural, khalil2017learning, nazari2018reinforcement, kool2018attention, kwon2020pomo, park2021schedulenet, cao2021dan, guo2019solving, wu2019learning, wu2021learning, falkner2020learning, chen2019learning} exhibit promising performances that are comparable to existing solvers as they learn solvers from the problem-solving simulations. \citep{bello2016neural, nazari2018reinforcement, kool2018attention,guo2019solving} utilize an encoder-decoder structure to generate routing schedules sequentially, while \citep{park2021schedulenet, khalil2017learning} use graph-based embedding to determine the next assignment action. However, RL approaches often requires the problem-specific Markov decision process and network design. NCE does not require the simulation of the entire problem-solving. Instead, it only requires computing the swapping operation (i.e., the results of CE). This property allows NCE to be trained easily to solve various routing problems with one scheme.


\paragraph{Neural network-based (meta) heuristic approach} Combining machine learning (ML) components with existing (meta) heuristics shows strong empirical performances when solving VRPs \cite{hottung2019neural, xin2021neurolkh, li2021learning, lu2019learning, da2021learning, kool2021deep}. They often employ ML to learn to solve NP-hard sub-problems of VRPs, which are difficult. For example, L2D \cite{li2021learning} learns to predict the objective value of CVRP, NLNS \cite{hottung2019neural} learns a TSP solver when solving VRPs and DPDP \cite{kool2021deep} learns to boost dynamic programming algorithms. To learn such solvers, these methods apply SL or RL. Instead, NCE learns the fundamental operator of meta-heuristics rather than predict or generate a solution. Hence, NCE that is trained on FMDVRP generalizes well to the special cases of FMDVRP. Furthermore, the training data for NCE can be prepared effortlessly.

\section{Experiments}
\label{sec:experiments}
This section provides the experiment results that validat the effectiveness of the proposed NCE in solving FMDVRP and the various VRPs. To train $f_\theta(\cdot)$, we use the input $(\tau1, \tau2, a1, a2)$ and output $y^*$ pairs obtained from 50,000 random FMDVRP instances. The details regarding the train data generation are described in \cref{appendix:train-detail}. The cost decrement model $f_\theta(\cdot)$ is parametrized by the GNN that contains the five attentive embedding layers. The details of the $f_\theta(\cdot)$ architecture and the computing infrastructure used to train $f_\theta(\cdot)$ are discussed in \cref{appendix:train-detail}. 

We emphasize that we use a single $f_\theta(\cdot)$ that is trained using FMDVRP for all experiments. We found that $f_\theta(\cdot)$ effectively solves the three special cases (i.e., MDVRP, mTSP, and CVRP) without retraining, proving the effectiveness of NCE as an universal operator for VRPs.

\subsection{FMDVRP experiments}
\label{subsec:FMDVRP-results}

\begin{table*}

\caption{\textbf{FMDVRP results} (small-sized instances)}
\vskip 0.1in
\label{table:fmdvrp_small}
\centering
\scriptsize
\begin{tabular}{crcccccc}
\toprule
$N_c$,$N_d$ & $N_v (\rightarrow)$ & \multicolumn{3}{c}{\textit{2}} & \multicolumn{3}{c}{\textit{3}} \\
$(\downarrow)$ & Method & Cost & Gap$(\%)$ & Time(sec.) & Cost & Gap$(\%)$ & Time(sec.) \\
\cmidrule(lr){2-2}\cmidrule(lr){3-5} \cmidrule(lr){6-8}

\multirow{3}{*}{(7,2)} 
& CPLEX & \textbf{1.543} &0.00 &0.31  & \textbf{1.363} &0.00  & 0.83   \\
& OR-tools & 1.596 & 3.43 &0.01  & 1.380 &1.25  & 0.01   \\
& CE & 1.546 & 0.02 &0.04  & 1.364 &0.01  & 0.03   \\
\cmidrule(lr){2-2}\cmidrule(lr){3-5} \cmidrule(lr){6-8}
& NCE & 1.546 & 0.02 & 0.10  & 1.365 & 0.01  & 0.12 \\
\midrule
 & $N_v (\rightarrow)$ & \multicolumn{3}{c}{\textit{2}} & \multicolumn{3}{c}{\textit{3}} \\
& Method & Cost & Gap$(\%)$ & Time(sec.) & Cost & Gap$(\%)$ & Time(sec.) \\
\cmidrule(lr){2-2}\cmidrule(lr){3-5} \cmidrule(lr){6-8}

\multirow{3}{*}{(10,2)} & CPLEX & \textbf{1.745} &0.00 & 9.29 & \textbf{1.488} &0.00  & 63.00   \\
& OR-tools & 1.820 & 4.30&0.02  & 1.521 &2.22  &  0.02 \\
& CE & 1.749 & 0.02 &0.07  & 1.493 &0.03  & 0.06   \\
\cmidrule(lr){2-2}\cmidrule(lr){3-5} \cmidrule(lr){6-8}
& NCE & 1.749 &0.02  & 0.13  & 1.493 & 0.03  & 0.16 \\
\bottomrule
\end{tabular}
\vskip -0.1in
\end{table*}

We evaluate the performance of NCE in solving various sizes of FMDVRP. We consider 100 random FMDVRP instances for each problem size $(N_c, N_d, N_v)$, where $N_c, N_d, N_v$ are the number of cities, depots, and vehicles, respectively. We provide the average makespan and computation time for the 100 instances. For small-sized problems ($N_c \leq 10$), we employ CPLEX \cite{cplex2009v12} (an exact method), OR-tools \cite{ortools}, and CE (full search) as the baselines. For the larger-sized problems, we exclude CPLEX from the baselines due to its limited scalability. To the best of our knowledge, our method is the first neural approach to solve FMDVRP; hence, we omit the neural baselines for FMDVRP. However, we include the neural baselines for mTSP and CVRP.

\begin{table*}
\caption{\textbf{FMDVRP results} (large-sized instances)}
\vskip 0.1in
\label{table:fmdvrp_larger}
\centering
\scriptsize
\begin{tabular}{crccccccccc}
\toprule
$N_c,N_d$ & $N_v(\rightarrow)$  & \multicolumn{3}{c}{\textit{3}}    & \multicolumn{3}{c}{\textit{5  }}     & \multicolumn{3}{c}{\textit{7}}        \\
$(\downarrow)$ & Method & Cost & Gap$(\%)$ & Time(sec.) & Cost & Gap$(\%)$ & Time(sec.) & Cost & Gap$(\%)$& Time(sec.) \\

\cmidrule(lr){2-2} \cmidrule(lr){3-5} \cmidrule(lr){6-8} \cmidrule(lr){9-11}

\multirow{3}{*}{(50,6)} 
& OR-tools & 2.39 &15.46 &2.20  & 1.56 &10.64 &2.44  & 1.27 &6.72  &2.58  \\
& CE &\textbf{2.07} &0.00 &21.06  & 1.41 &0.00 &9.09  & 1.19 &0.00  &5.37 \\
\cmidrule(lr){2-2} \cmidrule(lr){3-5} \cmidrule(lr){6-8} \cmidrule(lr){9-11}
& NCE &  2.08 &0.48 & 1.26 & \textbf{1.40} &-0.71  &1.82  & \textbf{1.19} &0.00  &2.23 \\

\midrule

& $N_v(\rightarrow)$ & \multicolumn{3}{c}{\textit{5}}    & \multicolumn{3}{c}{\textit{7  }}     & \multicolumn{3}{c}{\textit{10}}        \\

& Method & Cost & Gap$(\%)$ & Time(sec.) & Cost & Gap$(\%)$ & Time(sec.) & Cost & Gap$(\%)$& Time(sec.) \\
\cmidrule(lr){2-2} \cmidrule(lr){3-5} \cmidrule(lr){6-8} \cmidrule(lr){9-11}
\multirow{3}{*}{(100,8)}
& OR-tools & 2.00 &14.94 &30.46 & 1.51 &12.69 & 32.25 & 1.20 & 10.09 & 34.38\\
& CE & \textbf{1.74} &0.00  & 218.46&1.34 &0.00  &128.40  & 1.09 &0.00  &78.56  \\
\cmidrule(lr){2-2} \cmidrule(lr){3-5} \cmidrule(lr){6-8} \cmidrule(lr){9-11}
& NCE &  1.75 & 0.57 & 6.41  & \textbf{1.34} & 0.00  &9.54  & \textbf{1.09} & 0.00 & 13.34 \\
\bottomrule
\end{tabular}
\vskip -0.1in
\end{table*}

\cref{table:fmdvrp_small} shows the performances of NCE on the small-sized problems. NCE achieve similar makespans with CPLEX (optimal solution) within significantly lower computation times. NCE outperforms OR-tools in terms of makespan but has longer computation time; however, the computation time for NCE will be much lower than that of OR-tools when the problem size becomes bigger. It is noteworthy that NCE exhibits larger computation time than CE as the forward-propagation cost of GNN is larger than exhaustive search for small problems.

\cref{table:fmdvrp_larger} shows the performances of NCE on the large-sized problems. Applying CPLEX for large FMDVRPs is infeasible, so we exclude it from the baselines. Instead, the CE serves as an oracle to compute the makespans. For all cases, NCE has a near-zero gap compared to CE. This validates that NCE successfully amortizes the search operations of CE with significantly lower computation times. In addition, NCE consistently outperforms OR-tools for both the makespan and computational time. The performance gap between NCE and OR-tools becomes more significant as $N_c/N_v$ becomes large (i.e., each tour length becomes longer).

\paragraph{MDVRP results} We also apply the NCE with the $f_\theta$ that is trained on FMDVRP to solve MDVRP. As shown \cref{table:mdvrp_small,table:mdvrp_larger} in \cref{appendix:mdvrp}, NCE shows leading performance and is faster than the baselines similar to the FMDVRP experiments.


\subsection{mTSP experiments}
\label{subsec:mTSP-results}

\begin{table*}
\caption{\textbf{Average makespans of the random mTSPs:} DAN and ScehduleNet results are taken from the original papers, $\dagger$ Computational time of DAN is measured with the Nvidia RTX 3090.}
\vskip 0.1in
\label{table:mtsp}
\centering
\scriptsize
\begin{tabular}{rrccccccccc}
\toprule
$N_c$ & $N_v(\rightarrow$) & \multicolumn{3}{c}{\textit{5}} & \multicolumn{3}{c}{\textit{7  }} & \multicolumn{3}{c}{\textit{10}} \\
($\downarrow$) & Method & Cost & Gap$(\%)$ & Time(sec.) & Cost & Gap$(\%)$ & Time(sec.) & Cost &  Gap$(\%)$& Time(sec.)    \\
\cmidrule(lr){2-2} \cmidrule(lr){3-5} \cmidrule(lr){6-8} \cmidrule(lr){9-11}
\multirow{8}{*}{50} 
&LKH-3 & \textbf{2.00} &0.00 &187.46  &  \textbf{1.95} &0.00 &249.31  & 1.91 &0.00  &170.20  \\
&OR-tools & 2.04 & 2.00 & 3.24  & 1.96 &0.51 &3.75  & 1.91 &0.00  &3.67  \\
\cmidrule(lr){2-2} \cmidrule(lr){3-5} \cmidrule(lr){6-8} \cmidrule(lr){9-11}
&DAN& 2.29 & 14.50& 0.25$^\dagger$ & 2.11 &8.21 & 0.26$^\dagger$ & 2.03 & 6.28 & 0.30$^\dagger$\\
&ScheuduleNet & 2.17 &8.50 & 1.60 & 2.07 &6.15 & 1.67 & 1.98 & 3.66 &1.90 \\
\cmidrule(lr){2-2} \cmidrule(lr){3-5} \cmidrule(lr){6-8} \cmidrule(lr){9-11}
& NCE & 2.02 &1.00 & 2.25 & \ 1.96 &0.51 & 2.44 & \textbf{1.91} & 0.00 & 3.38 \\
& NCE-mTSP & 2.02 &1.00 & 2.48 & \ 1.96 &0.51 & 2.50 & \textbf{1.91} & 0.00 & 3.44 \\
\midrule
\multirow{10}{*}{\textit{100}} & $N_v(\rightarrow$) & \multicolumn{3}{c}{\textit{5}} & \multicolumn{3}{c}{\textit{10 }} & \multicolumn{3}{c}{\textit{15}} \\
& Method & Cost & Gap$(\%)$ & Time(sec.) & Cost & Gap$(\%)$ & Time(sec.) & Cost & Gap$(\%)$& Time(sec.) \\
\cmidrule(lr){2-2} \cmidrule(lr){3-5} \cmidrule(lr){6-8} \cmidrule(lr){9-11}
&LKH-3 & \textbf{2.20} &0.00 &262.85  & 1.97 &0.00 &474.78  & 1.98 &0.00  &378.90  \\
&OR-tools & 2.41 &9.55 &35.47 & 2.03 &3.05 &45.40  & 2.03 & 2.53 &48.86  \\
\cmidrule(lr){2-2} \cmidrule(lr){3-5} \cmidrule(lr){6-8} \cmidrule(lr){9-11}
&DAN& 2.72 &23.64 & 0.43$^\dagger$ & 2.17 &10.15 & 0.48$^\dagger$  & 2.09& 5.56 & 0.58$^\dagger$ \\
&ScheuduleNet & 2.59 & 17.73&14.84  & 2.13 &8.12 &16.22  & 2.07 &4.55  &20.02  \\
\cmidrule(lr){2-2} \cmidrule(lr){3-5} \cmidrule(lr){6-8} \cmidrule(lr){9-11}
& NCE & 2.25 &2.27 & 16.01 & 1.98 &0.51 & 12.22 & \textbf{1.98} & 0.00 & 24.08 \\
& NCE-mTSP & 2.24 &1.82 & 16.36 & \textbf{1.97} &0.00 & 13.00 & \textbf{1.98} & 0.00 & 23.37 \\
\midrule

\multirow{10}{*}{\textit{200}} & $N_v(\rightarrow)$  & \multicolumn{3}{c}{\textit{10}}    & \multicolumn{3}{c}{\textit{15 }} & \multicolumn{3}{c}{\textit{20}}        \\
& Method & Cost & Gap$(\%)$ & Time(sec.) & Cost & Gap$(\%)$ & Time(sec.) & Cost & Gap$(\%)$& Time(sec.) \\
\cmidrule(lr){2-2} \cmidrule(lr){3-5} \cmidrule(lr){6-8} \cmidrule(lr){9-11}

&LKH-3 & \textbf{2.04} &0.00 & 1224.40  & 2,00  &0.00 & 1147.13 &  \textbf{1.97} &0.00  & 908.14 \\
&OR-tools &2.33 & 14.22 &675.79 & 2.33 & 16.50&604.31  & 2.37 & 20.30 &649.17 \\
\cmidrule(lr){2-2} \cmidrule(lr){3-5} \cmidrule(lr){6-8} \cmidrule(lr){9-11}

&DAN& 2.40 & 17.65 &0.93$^\dagger$  & 2.20 & 10.00& 0.98$^\dagger$ & 2.15& 9.14 & 1.07$^\dagger$ \\
&ScheuduleNet & 2.45 & 20.10&193.41  & 2.24 &12.00 &213.07  & 2.17 &10.15  &225.50  \\
\cmidrule(lr){2-2} \cmidrule(lr){3-5} \cmidrule(lr){6-8} \cmidrule(lr){9-11}
& NCE & 2.06 &0.98 & 83.82 &\textbf{2.00} &0.00 & 72.32 & 2.02 & 2.54 & 118.70 \\
& NCE-mTSP & 2.06 &0.98 & 84.96 &\textbf{2.00} &0.00 & 84.28 & 2.02 & 2.54 & 108.91 \\
\bottomrule
\end{tabular}
\vskip -0.1in
\end{table*}

\begin{table*}[t]
\centering
\renewcommand{\arraystretch}{1.3} 
\caption{\textbf{mTSPLib results}: CPLEX results with $*$ are optimal solutions. Otherwise, the known-best upper bound of CPLEX results are reported. The results of other baselines are taken from \cite{park2021schedulenet}.}

\vskip 0.15in
\resizebox{\textwidth}{!}{
\begin{tabular}{rccccccccccccccccc}
\toprule
$N_c (\rightarrow)$ & \multicolumn{4}{c}{\textit{Eil51}}    & \multicolumn{4}{c}{\textit{Berlin52}}     & \multicolumn{4}{c}{\textit{Eil76}}    & \multicolumn{4}{c}{\textit{Rat99}} &     \\
\cmidrule(lr){2-5}
\cmidrule(lr){6-9}
\cmidrule(lr){10-13}
\cmidrule(lr){14-17}
$N_v (\rightarrow)$         & 2     & 3     & 5     & 7     & 2      & 3      & 5      & 7      & 2     & 3     & 5     & 7     & 2     & 3     & 5     & 7     & Gap         \\
\cmidrule(lr){1-1}
\cmidrule(lr){2-5}
\cmidrule(lr){6-9}
\cmidrule(lr){10-13}
\cmidrule(lr){14-17}
\cmidrule(lr){18-18}
CPLEX & 222.7$^*$ & 159.6 & 124.0 & 112.1 & 4110 & 3244 & 2441 & 2441 & 280.9$^*$ & 197.3 & 150.3 & 139.6 & 728.8 & 587.2 & 469.3 & 443.9 & 1.00 \\
\cmidrule(lr){1-1}
\cmidrule(lr){2-5}
\cmidrule(lr){6-9}
\cmidrule(lr){10-13}
\cmidrule(lr){14-17}
\cmidrule(lr){18-18}
LKH-3 & \textbf{222.7} & \textbf{159.6} & 124.0 & 112.1 & 4110 & 3244 & \textbf{2441} & 2441 & \textbf{280.9} & \textbf{197.3} & 150.3 & 139.6 & 728.8 & 587.2 & 469.3 & 443.9 & 1.00 \\ 
OR-Tools & 243.0 & 170.1 & 127.5 & 112.1 & 4665 & 3311 & 2482 & 2441 & 318.0 & 212.4 & 143.4 & 128.3 & 762.2 & 552.1 & 473.7 & 442.5 & 1.03                     \\ 
ScheduleNet & 263.9 & 200.5 & 131.7 & 116.9 & 4826 & 3644 & 2758 & 2515 & 330.2 & 228.8 & 163.9 & 144.4 & 843.8 & 691.8 & 524.3 & 480.8 & 1.13                     \\
ScheduleNet (s.64) & 239.3 & 173.5 & 125.8 & 112.2 & 4592 & 3276 & 2517 & 2441 & 317.7 & 220.8 & 153.8 & 131.7 & 781.2 & 627.1 & 502.3 & 464.4 & 1.05                     \\
DAN & 274.2 & 178.9 & 158.6 & 118.1 & 5226 & 4278 & 2759 & 2697 & 361.1 & 251.5 & 170.9 & 148.5 & 930.8 & 674.1 & 504.0 & 466.4 & 1.18                     \\
DAN (s.64) & 252.9 & 178.9 & 128.2 & 114.3 & 5098 & 3456 & 2677 & 2495 & 336.7 & 228.1 & 157.9 & 134.5 & 966.5 & 697.7 & 495.6 & 462.0 & 1.11                     \\
\cmidrule(lr){1-1}
\cmidrule(lr){2-5}
\cmidrule(lr){6-9}
\cmidrule(lr){10-13}
\cmidrule(lr){14-17}
\cmidrule(lr){18-18}
NCE & 235.0 & 170.3 & 121.6 & \textbf{112.1} & \textbf{4110} & 3274 & 2660 & \textbf{2441} & 285.5 & 211.0 & \textbf{144.6} & \textbf{127.6} & 695.8 & \textbf{527.8} & \textbf{458.6} & \textbf{441.6} &   1.00                   \\ 
NCE-mTSP     & 226.1 & 166.3 &\textbf{119.9} & \textbf{112.1}& 4128 &\textbf{3191} & 2474 & \textbf{2441} & 282.1 & 197.5 & 147.2 & \textbf{127.6} & \textbf{666.0} & 533.2 & 462.2 & 443.9 & \textbf{0.98}                   \\ 
\bottomrule
\end{tabular}
}
\vspace{-0.3cm}
\label{table:mtsp-benchmark}
\end{table*}

We evaluate NCE when solving mTSP. We provide the average performance of 100 instances for each $(N_c, N_v)$ pair. For the baselines, we consider two meta-heuristics (LKH-3 \cite{helsgaun2017extension}, which is known as the one of the best mTSP heuristics, and OR-tools) and two neural baselines (ScheduleNet \cite{2021scnet} and DAN \cite{cao2021dan}).

As shown in \cref{table:mtsp}, NCE achieves similar performance with LKH-3 within significantly shorter computational time. It is noteworthy that LKH-3 employs mTSP-specific heuristics on top of LKH heuristics, while NCE do not employ any mTSP-specific structures. To validate the effect of task-specific information on NCE, we train NCE with mTSP data (NCE-mTSP) and solve mTSP. The performances of NCE and NCE-mTSP are almost identical, which indicates that NCE is highly generalizable. In addition, NCE consistently outperforms the neural baseline. We further apply NCE to solve mTSPLib \cite{mTSPLib}, which comprise of mTSP instances from real cities. As reported in \cref{table:mtsp-benchmark}, NCEs achieves the best results as compared to the baselines.

\subsection{CVRP experiments}
\label{subsec:CVRP-results}

We evaluate NCE when solving capacitated VRP (CVRP), a canonical VRP problem that has additional capacity constraints. Even though training $f_\theta(\cdot)$ is done without the consideration of the capacity constraints, we can easily enforce such constraints without retraining $f_\theta(\cdot)$ by adjusting the searching range as follows:
\begin{align}
    (b_1, b_2)\leftarrow\argmax_{b_1, b_2\in S_c}\left(C(\text{\fontfamily{lmtt}\selectfont CROSS}((a_1,b_1,a_2,b_2; \tau_1, \tau_2)))-C(\tau_1, \tau_2)\right),
\end{align}
where the searching range $S_c$ is a set of nodes that satisfies the capacity constraints. As shown in \cref{table:cvrp_benchmark}, NCE is on par with or outperforms other neural baselines, which again proves the effectiveness of NCE as an universal operator. 

\begin{table*}[t]
\caption{\textbf{CVRP benchmark results:} $(s.n)$ indicates the best results of $n$ sampling, $(i.n)$ indicates the best results after $n$ improvement steps, and $\dagger$ the computation times of neural baselines are measured with GPU. The run times of the neural baselines are taken from \cite{kim2021learning}.}
\vskip 0.1in
\label{table:cvrp_benchmark}
\centering
\scriptsize
\begin{tabular}{lccccccccc}
\toprule
& \multicolumn{3}{c}{\textit{CVRP20}} & \multicolumn{3}{c}{\textit{CVRP50}} & \multicolumn{3}{c}{\textit{CVRP100}} \\

Method& Cost & Gap$(\%)$ &Time(sec.) & Cost & Gap$(\%)$ &Time(sec.) & Cost & Gap$(\%)$ &Time(sec.)    \\
\cmidrule(lr){1-1} \cmidrule(lr){2-4} \cmidrule(lr){5-7} \cmidrule(lr){8-10}
LKH-3&  6.14 & 0.00&  0.72 & \textbf{10.38} &0.00  & 2.52  & \textbf{15.65} &0.00  &4.68 \\
OR-Tools &   6.43&4.72 &0.01  &11.31 &8.17 & 0.05  & 17.16 &10.29 &0.23 \\
\cmidrule(lr){1-1} \cmidrule(lr){2-4} \cmidrule(lr){5-7} \cmidrule(lr){8-10}
RL$^\dagger (s.10)$ \cite{nazari2018reinforcement} & 6.40  &4.23 & 0.16 &11.15  &7.46  &0.23  &16.96  & 8.39 &0.45 \\
AM$^\dagger (s.1280)$ \cite{kool2018attention} & 6.25  &1.79 & 0.05 &10.62  &2.40  &0.14  &16.23  & 3.72 &0.34 \\
MDAM$^\dagger (s.50)$ \cite{xin2021multi}& 6.14  &0.00 & 0.03  &10.48  &0.96  & 0.09 & 15.99 &2.17   &0.32 \\
POMO$^\dagger (s.8)$ \cite{kwon2020pomo}& 6.14  &0.00 & 0.01 &10.42  &0.35  & 0.01 & 15.73 &0.43  &0.01 \\

\cmidrule(lr){1-1} \cmidrule(lr){2-4} \cmidrule(lr){5-7} \cmidrule(lr){8-10}

NLNS$^\dagger (i.1280)$ \cite{hottung2019neural}& 6.19  & 0.81& 1.00&10.54 & 1.54 &1.63 &16.00  &2.24 & 2.18 \\
AM + LCP$^\dagger (s.1280)$ \cite{kim2021learning} & 6.16  &0.33 & 0.09 &10.54  &1.54  &0.20  &16.03  & 2.43 &0.45 \\
\cmidrule(lr){1-1} \cmidrule(lr){2-4} \cmidrule(lr){5-7} \cmidrule(lr){8-10}
NCE & 6.22  & 1.30 &0.73  & 10.72 & 3.17 &3.14  &16.33  & 4.35 &13.60  \\
NCE (s.10) & \textbf{6.14} & 0.00&1.79  & 10.49 & 1.06  &8.04  & 16.00 & 2.24 & 33.85 \\
\bottomrule
\end{tabular}
\vskip -0.1in
\end{table*}

\subsection{Ablation studies}
\label{subsec:Parametric study}
We evaluate the effects of the hyperparameters on NCE. The results are as follows:
\begin{itemize}[leftmargin=0.5cm]
\vspace{-0.25cm}
    \item \cref{appendix-subsub:candidate}: the performance of NCE converges when the number of candidate $K \geq 10$.
    \item \cref{appendix-subsub:intra-solver}: the performance of NCE is less sensitive to the selection of intra solvers.
    \item \cref{appendix-subsub:choosing-2-vehicle}: the performance of NCE is less sensitive to the selection of swapping tours. 
    \item \cref{appendix-subsub:perturbation}: the performance of NCE converges when the perturbation parameter $p \geq 5$.
\end{itemize}
\vspace{-0.25cm}

\section{Conclusion}

We propose Neuro CROSS exchange (NCE), a neural network-enhanced CE operator, to learn a fundamental and universal operator that can be used to solve the various types of practical VRPs. NCE learns to predict the best cost-decrements of the CE operation and utilizes the prediction to amortize the costly search process of CE. As a result, NCE reduces the search cost of CE from $\mathcal{O}(N^4)$ to $\mathcal{O}(N^2)$. Furthermore, the NCE operator can learn with data that are relatively easy to obtain, which reduces training difficulty. We validated that NCE can solve various VRPs without training for each specific problem, exhibiting strong empirical performances. 

Although NCE addresses more realistic VRPs (i.e., FMDVRP) than existing NCO solvers, NCE does not yet consider complex constraints such as pickup and delivery, and time windows. Our future research will focus on solving more complex VRP by considering such various constraints during the NCE operation.

\bibliography{references}


\appendix
\renewcommand \thepart{}
\renewcommand \partname{}

\newpage
\rule[0pt]{\columnwidth}{3pt}
\begin{center}
    \huge{\bf{Neuro CROSS exchange} \\
    \emph{Supplementary Material}}
\end{center}
\vspace*{3mm}
\rule[0pt]{\columnwidth}{1pt}
\vspace*{-.5in}

\appendix
\addcontentsline{toc}{section}{}
\part{}
\parttoc

\renewcommand{\theequation}{A.\arabic{equation}}

\setcounter{equation}{0}

\clearpage

\section{MILP formulations for min-max Routing Problems}

This section provides the mixed integer linear programming (MILP) formulations of mTSP, MDVRP, and FMDVRP.

\subsection{mTSP}
\label{sub:mTSP_form}
mTSP is a multi-vehicle extension of the traveling salesman problem (TSP). mTSP comprises the set of the nodes (i.e., cities) and the depot $V$, the set of vehicles $K$, and the set of depot $S$. We define $d_{ij}$ as the cost (or travel time) between node $i$ and $j$, and the decision variable $x_{ijk}$ which denotes whether the edge between node $i$ and $j$ are taken by vehicle $k$. Following the convention, we consider mTSP with $|S|=1$. The MILP formulation of mTSP is given as follows:
\begin{align}
\label{mTSP_form}
&\text{minimize}&& Q&&\\
&\text{subject to.}& & \sum_{i \in V}\sum_{j \in V} d_{ij}x_{ijk} \leq Q, && \forall k\in K : i\neq j, &\label{eq:mtsp:const1} \\
&& & \sum_{j \in V i \neq j} x_{ijk} = 1, && \forall k \in K, \forall i \in S,& \label{eq:mtsp:const2}\\
&& &  \sum_{i \in V j\neq i} \sum_{k \in T} x_{ijk} = 1, &&\forall j \in V \setminus S&\label{eq:mtsp:const3} \\
&& & \sum_{i \in V i\neq j} x_{ijk}  -  \sum_{h \in V h\neq j} x_{jhk} = 0, &&\forall j \in V \setminus S&\label{eq:mtsp:const4} \\
&& & u_{ik} - u_{jk} + |V| x_{ijk} \leq |V|-1, && \forall k \in K, j \in V \setminus S : i \neq j,&\label{eq:mtsp:const5} \\
&& & 0 \leq u_{ik} \leq |V|-1, && \forall k \in K, i \in V \setminus S &\label{eq:mtsp:const6} \\
&& & x_{ijk} \in \{0,1\}, && \forall k \in K, \forall i,j \in V,& \\
&& & u_{ik} \in \mathbb{Z}, && \forall k \in K, i \in V & 
\end{align}
where $Q$ denotes the longest traveling distance among multiple vehicles. (i.e., makespan), \cref{eq:mtsp:const2} indicates the vehicles start at the depot, \cref{eq:mtsp:const3}  indicates all cities are visited, \cref{eq:mtsp:const4} indicates the balance equation for all cities, \cref{eq:mtsp:const5} and \cref{eq:mtsp:const6} indicate the sub-tour eliminations.  

\newpage
\subsection{MDVRP}
\label{sub:MDVRP_form}
Multi-depot VRP is a multi-depot extension of mTSP (\cref{sub:mTSP_form}) where each vehicle starts from its own designated depot and returns to the depot. We extend the MILP formulation of mTSP to define the MILP formulation of MDVRP. On top of the mTSP formulation, we define $K_i$, which indicates the set of vehicles assigned to the depot $i$.
\begin{align}
\label{mDVRP_form}
&\text{minimize}&& Q&&\\
&\text{subject to.}& & \sum_{i \in V}\sum_{j \in V}   d_{ij}x_{ijk} \leq Q,  && \forall k\in K : i\neq j, & \\
&& & \sum_{j \in V j\neq i} \sum_{k \in T} x_{ijk} = 1, &&\forall i \in V \setminus S& \\
&& &  \sum_{i \in V j\neq i} \sum_{k \in T} x_{ijk} = 1, && \forall j \in V \setminus S& \\
&& & \sum_{i \in V } x_{ijk}  -  \sum_{h \in V } x_{jhk} = 0, && \forall j \in V \setminus S,    \forall k \in K& \\
&& & u_{ik} - u_{jk} + |V| x_{ijk} \leq |V|-1, && \forall k \in K, j \in V \setminus S : i \neq j,& \\
&& & 0 \leq u_{ik} \leq |V|-1, && \forall k \in K, i \in V \setminus S & \\
&& & x_{ijk} \in \{0,1\}, && \forall k \in K, \forall i,j \in V,& \\
&& & u_{ik} \in \mathbb{Z}, && \forall k \in K, i \in V &  \\
&& & \sum_{j \in V \setminus S}  x_{ijk} \leq 1, && \forall k \in K_{i} ,\forall i \in S &\label{eq:mdvrp:const5} \\
&& & \sum_{i \in V \setminus S}  x_{ijk} \leq 1, && \forall k \in K_{j} ,\forall j \in S & \label{eq:mdvrp:const6}
\end{align}
where \cref{eq:mdvrp:const5} and \cref{eq:mdvrp:const6} indicate that each vehicle starts and returns its own depot at most once.

\newpage

\subsection{FMDVRP}
\label{appendix:fmdvrp}
Flexible MDVRP is an extension of MDVRP, allowing the vehicle to return to any depot. We extend the MDVRP formulation (\cref{sub:MDVRP_form}) to define the FMDVRP formulation.
To account for the flexibility of depot returning, we introduce a dummy node for all depots; therefore, a depot is modeled with a start and return depot. We define $S_1$ and $S_2$ as the set of start and return depots and $s_k$ as the start node of the vehicle $k$.
\begin{align}
\label{FMDVRP_form}
&\text{minimize}&& Q&&\\
&\text{subject to.}& & \sum_{i \in V}\sum_{j \in V}   d_{ij}x_{ijk} \leq Q,  && \forall k\in K : i\neq j, &\\
&& & \sum_{j \in V j\neq i} \sum_{k \in T} x_{ijk} = 1, &&\forall i \in V \setminus S& \\
&& &  \sum_{i \in V j\neq i} \sum_{k \in T} x_{ijk} = 1, && \forall j \in V \setminus S&\\
&& & \sum_{i \in V } x_{ijk}  -  \sum_{h \in V } x_{jhk} = 0, && \forall j \in V \setminus S,     \forall k \in K& \\
&& & u_{ik} - u_{jk} + |V| x_{ijk} \leq |V|-1, && \forall k \in K, j \in V \setminus S : i \neq j,& \\
&& & 0 \leq u_{ik} \leq |V|-1, && \forall k \in K, i \in V \setminus S & \\
&& & x_{ijk} \in \{0,1\}, && \forall k \in K, \forall i,j \in V,& \\
&& & u_{ik} \in \mathbb{Z}, && \forall k \in K, i \in V &  \\
&& & \sum_{j \in V \setminus S}  x_{s_{k}jk} = 1, && \forall k \in K  \label{eq:fmdvrp:const1}\\
&& & \sum_{j \in V \setminus S}  x_{ijk} = 0, && \forall k \in K , \forall i \in S \setminus s_{k} \label{eq:fmdvrp:const2}\\
&& & \sum_{j \in V \setminus S}  x_{ijk} \leq 1, && \forall k \in K_{i} ,\forall i \in S1 &\label{eq:fmdvrp:const3} \\
&& & \sum_{i \in V \setminus S}  x_{ijk} \leq 1, && \forall k \in K_{j} ,\forall j \in S2 & \label{eq:fmdvrp:const4}\\
&& & \sum_{j \in V \setminus S}  x_{ijk} = 0, && \forall k \in K , \forall i \in S2 &\label{eq:fmdvrp:const5}\\
&& & \sum_{j \in V \setminus S}  x_{ijk} = 0, && \forall k \in K , \forall i \in S1 &\label{eq:fmdvrp:const6}\\
&& & \sum_{i \in S1} \sum_{j \in V \setminus S}  x_{ijk} = \sum_{i \in V \setminus S} \sum_{j \in S2}  x_{ijk}, && \forall k \in K && \label{eq:fmdvrp:const7}
\end{align}
where \cref{eq:fmdvrp:const1,eq:fmdvrp:const2} indicate each vehicle starts at its own depot. \cref{eq:fmdvrp:const3,eq:fmdvrp:const4,eq:fmdvrp:const5,eq:fmdvrp:const6} indicate start and return depots constraints. \cref{eq:fmdvrp:const7} indicates the balance equation of the start and return depots.
\newpage
\section{MDVRP result}

\label{appendix:mdvrp}

In this section, we provide the experiment results of MDVRP. We apply NCE with the $f_\theta$ trained on FMDVRP instances to solve MDVRP. For each $(N_c, N_d, N_v)$ pair, we measure the average makespan of 100 instances. We provide the MDVRP results in \cref{table:mdvrp_small,table:mdvrp_larger}. Similar to the FMDVRP experiments, NCE shows leading performance while faster than the baselines. From the results, we can conclude that the learned $f_\theta$ is transferable to the different problem sets. This phenomenon is rare in many ML-based approaches. It again highlights the effectiveness of learning fundamental operators (i.e., learn to what should be cross exchanged) in solving VRP families.

\begin{table*}[h!]
\caption{\textbf{MDVRP results} (small size instances)}
\vskip 0.1in
\label{table:mdvrp_small}
\centering
\scriptsize
\begin{tabular}{crcccccc}
\toprule
$N_c$,$N_d$ & $N_v (\rightarrow)$ & \multicolumn{3}{c}{\textit{2}} & \multicolumn{3}{c}{\textit{3}} \\
$(\downarrow)$ & Method & Cost & Gap$(\%)$ & Time(sec.) & Cost & Gap$(\%)$ & Time(sec.) \\
\cmidrule(lr){2-2}\cmidrule(lr){3-5} \cmidrule(lr){6-8}

\multirow{3}{*}{(7,2)} 
& CPLEX & \textbf{1.626} &0.00 &0.32  & \textbf{1.417} &0.00  & 0.54   \\
& OR-tools & 1.704 & 4.80 &0.01  & 1.433 &1.13  & 0.01   \\
& CE & \textbf{1.626} & 0.00 &0.05  & 1.418 &0.01  & 0.04   \\
\cmidrule(lr){2-2}\cmidrule(lr){3-5} \cmidrule(lr){6-8}
& NCE &  \textbf{1.626} & 0.00 & 0.13  & 1.418 & 0.01  & 0.16 \\
\midrule
 & $N_v (\rightarrow)$ & \multicolumn{3}{c}{\textit{2}} & \multicolumn{3}{c}{\textit{3}} \\
& Method & Cost & Gap$(\%)$ & Time(sec.) & Cost & Gap$(\%)$ & Time(sec.) \\
\cmidrule(lr){2-2}\cmidrule(lr){3-5} \cmidrule(lr){6-8}

\multirow{3}{*}{(10,2)} & CPLEX & \textbf{1.829} &0.00 & 7.90 & \textbf{1.554} &0.00  & 33.17   \\
& OR-tools & 1.926 & 5.30&0.02  & 1.590 &2.32  &  0.02 \\
& CE & \textbf{1.829}& 0.00&0.09  & 1.558 &0.03  &  0.08 \\
\cmidrule(lr){2-2}\cmidrule(lr){3-5} \cmidrule(lr){6-8}
& NCE & \textbf{1.829} &0.00  & 0.17  & 1.555 & 0.01  & 0.20 \\
\bottomrule
\end{tabular}
\end{table*}

\begin{table*}[h!]
\caption{\textbf{MDVRP results} (large size instances)}
\vskip 0.1in
\label{table:mdvrp_larger}
\centering
\scriptsize
\begin{tabular}{crccccccccc}
\toprule
$N_c,N_d$ & $N_v(\rightarrow)$  & \multicolumn{3}{c}{\textit{3}}    & \multicolumn{3}{c}{\textit{5  }}     & \multicolumn{3}{c}{\textit{7}}        \\
$(\downarrow)$ & Method & Cost & Gap$(\%)$ & Time(sec.) & Cost & Gap$(\%)$ & Time(sec.) & Cost & Gap$(\%)$& Time(sec.) \\

\cmidrule(lr){2-2} \cmidrule(lr){3-5} \cmidrule(lr){6-8} \cmidrule(lr){9-11}

\multirow{3}{*}{(50,6)} 
& OR-tools & 2.64 &17.33 &2.24  & 1.68 &9.80 &2.94  & 1.36 &6.25  &2.75  \\
& CE & 2.25 &0.00 &23.45  & 1.53 &0.00 &10.40  & 1.28 &0.00  &6.85 \\
\cmidrule(lr){2-2} \cmidrule(lr){3-5} \cmidrule(lr){6-8} \cmidrule(lr){9-11}
& NCE & \textbf{2.25} &0.00 & 2.08 & \textbf{1.53} &0.00 & 2.63 & \textbf{1.28} & 0.00 & 2.93 \\
\midrule

& $N_v(\rightarrow)$ & \multicolumn{3}{c}{\textit{5}}    & \multicolumn{3}{c}{\textit{7  }}     & \multicolumn{3}{c}{\textit{10}}        \\
& Method & Cost & Gap$(\%)$ & Time(sec.) & Cost & Gap$(\%)$ & Time(sec.) & Cost & Gap$(\%)$& Time(sec.) \\
\cmidrule(lr){2-2} \cmidrule(lr){3-5} \cmidrule(lr){6-8} \cmidrule(lr){9-11}
\multirow{3}{*}{(100,8)}
& OR-tools & 2.17 &17.30 &33.08 & 1.60 &11.89 & 36.45 & 1.29 & 9.32 & 37.54\\
& CE & \textbf{1.85} &0.00 &259.82  & 1.43 &0.00  &140.63  & 1.18 &0.00  &86.27  \\
\cmidrule(lr){2-2} \cmidrule(lr){3-5} \cmidrule(lr){6-8} \cmidrule(lr){9-11}
& NCE &  1.86 & 0.54 & 11.61  & \textbf{1.43} & 0.00  &11.96  & \textbf{1.18} & 0.00 & 15.70 \\
\bottomrule
\end{tabular}
\end{table*}


\newpage
\section{Ablation study}
\label{appendix:param}
In this section, we provide the results of the ablation studies. 
\subsection{Candidate set}
\label{appendix-subsub:candidate}

NCE constructed a search candidate set. To mitigate the prediction error of $f_{\theta}(\cdot)$ in finding the argmax $(a_1, a_2, b_1, b_2)$,  NCE search the top $K$ pairs of $(a_1, a_2)$ that have the largest $y^*$ out of all $(a_1,a_2)$ choices. We measured how the performance changes whenever the size of the candidate set $K$changes. As shown in \cref{table:number of candidate}, 
as the size of $K$ increases, the performance tends to increase slightly. When $K \geq 10$, the performance of NCE almost converges. Thus, we choose $K=10$ as the default hyperparameter of NCE.
\begin{table*}[h!]

\centering
\renewcommand{\arraystretch}{1.3}

\caption{\textbf{Effect of number of candidate}}

\vskip 0.15in
\resizebox{\textwidth}{!}{
\begin{tabular}{crcccccccccccccccc}
\midrule
$K$ & \multicolumn{2}{c}{\textit{1}}    & \multicolumn{2}{c}{\textit{2}}     & \multicolumn{2}{c}{\textit{3}}    & \multicolumn{2}{c}{\textit{5}}  & \multicolumn{2}{c}{\textit{7}}& \multicolumn{2}{c}{\textit{10}}& \multicolumn{2}{c}{\textit{20}}& \multicolumn{2}{c}{\textit{30}}  \\
\cmidrule(lr){1-1}\cmidrule(lr){2-3}\cmidrule(lr){4-5}\cmidrule(lr){6-7}\cmidrule(lr){8-9}\cmidrule(lr){10-11}\cmidrule(lr){12-13}\cmidrule(lr){14-15}\cmidrule(lr){16-17}
$N_c$,$N_d$,$N_v$        &  cost    & time     & cost     & time     & cost      & time      & cost      & time      & cost     & time     & cost    & time     & cost     & time     & cost    & time                               \\
\cmidrule(lr){1-1}\cmidrule(lr){2-3}\cmidrule(lr){4-5}\cmidrule(lr){6-7}\cmidrule(lr){8-9}\cmidrule(lr){10-11}\cmidrule(lr){12-13}\cmidrule(lr){14-15}\cmidrule(lr){16-17}
(30,3,2)       & 2.47 & 0.26 & 2.44 & 0.30 & 2.44 & 0.34 & 2.43 & 0.38 & 2.43 & 0.43 & 2.43 & 0.48 & 2.43 & 0.63 & 2.43 & 0.81                     \\ 
(30,3,3)       & 1.87 & 0.27 & 1.85 & 0.31 & 1.84 & 0.35 & 1.84 & 0.41 & 1.83 & 0.48 & 1.83 & 0.55 & 1.83 & 0.79 & 1.83 & 1.04                     \\ 
(30,3,5)       & 1.50 & 0.50 & 1.47 & 0.61 & 1.47 & 0.66 & 1.46 & 0.71 & 1.46 & 0.83 & 1.46 & 0.91 & 1.47 & 1.27 & 1.46 & 1.54                     \\ 

(50,3,3)    & 2.23 & 0.58 & 2.20 & 0.80 & 2.19 & 0.93 & 2.18 & 1.13 & 2.19 & 1.26 & 2.18 & 1.51 & 2.19 & 2.17 & 2.18 & 2.70                     \\ 
(50,3,5) & 1.67 & 0.86 & 1.63 & 1.12 & 1.62 & 1.34 & 1.61 & 1.56 & 1.61 & 1.81 & 1.61 & 2.17 & 1.61 & 3.14 & 1.61 & 4.07                     \\
(50,3,7) & 1.49 & 1.05 & 1.47 & 1.31 & 1.47 & 1.59 & 1.46 & 1.93 & 1.46 & 2.18 & 1.46 & 2.59 & 1.46 & 3.79 & 1.46 & 4.98                     \\

\bottomrule
\end{tabular}
}
\label{table:number of candidate}
\end{table*}
\newpage
\subsection{Intra-solver}
\label{appendix-subsub:intra-solver}

NCE repeatedly applies the inter-and intra-operation. In this view, the choice of intra-operation may affect the performance of NCE. In this subsection, we measured the performance of NCE according to intra-operation. We compare the results of NCE with Elkai, OR-tools, and 2-opt as the intra-operator. To solve TSP -- the task intra-operator has to solve --, Elkai, OR-tools, and 2-opt show the best, second best, and third best performances. As shown in \cref{table:intra-solver}, the performances of NCE are almost identical to the selection of an intra-operator. We validate that the effect of intra-operation choice is negligible to the performance.  
\begin{table*}[h!]
\centering
\renewcommand{\arraystretch}{1.3} 

\caption{\textbf{Effect of Intra TSP solver}}
\vskip 0.15in

\resizebox{\textwidth}{!}{
\begin{tabular}{crcccccccccccc}
\midrule
$N_c$,$N_d$,$N_v$   & \multicolumn{2}{c}{\textit{(30,3,2)}}    & \multicolumn{2}{c}{\textit{(30,3,3)}}     & \multicolumn{2}{c}{\textit{(30,3,5)}}    & \multicolumn{2}{c}{\textit{(50,3,3)}}  & \multicolumn{2}{c}{\textit{(50,3,5)}}& \multicolumn{2}{c}{\textit{(50,3,7)}}  \\
\cmidrule(lr){1-1}\cmidrule(lr){2-3}\cmidrule(lr){4-5}\cmidrule(lr){6-7}\cmidrule(lr){8-9}\cmidrule(lr){10-11}\cmidrule(lr){12-13}
Intra solver        &  cost    & time     & cost     & time     & cost      & time      & cost      & time      & cost     & time     & cost    & time                                   \\
\cmidrule(lr){1-1}\cmidrule(lr){2-3}\cmidrule(lr){4-5}\cmidrule(lr){6-7}\cmidrule(lr){8-9}\cmidrule(lr){10-11}\cmidrule(lr){12-13}
2-opt       & 2.46 & 0.23 & 1.83 & 0.33 & 1.47 & 0.53 & 2.22 & 0.72 & 1.62 & 1.24 & 1.46 & 1.58                     \\ 
OR-tools & 2.44 & 1.04 & 1.83 & 1.08 & 1.47 & 1.06 & 2.20 & 3.31 & 1.61 & 2.72 & 1.46 & 2.72                  \\
Elkai   & 2.43 & 0.41 & 1.83 & 0.55 & 1.46 & 0.69 & 2.18 & 1.55 & 1.61 & 2.17 & 1.46 & 2.13                    \\

\bottomrule
\end{tabular}
}
\label{table:intra-solver}
\end{table*}
\newpage
\subsection{Selecting two vehicles}
\label{appendix-subsub:choosing-2-vehicle}

NCE chooses two tours for improvement during the iterative process. 
To understand the effect of the tour selection strategy, we measured the performance of NCE according to tour selection. We compared NCE results in a max-min selection case and a random selection case (i.e., pick two tours randomly). As shown in \cref{table:choosing 2 vehicle}, the performances of NCE are almost identical to the tour selection strategy. Therefore, we validate that the effect of the tour selection strategy is negligible.
\begin{table*}[h!]
\centering
\renewcommand{\arraystretch}{1.3} 

\caption{\textbf{Effect of selecting two vehicles}}

\vskip 0.15in
\resizebox{\textwidth}{!}{
\begin{tabular}{crcccccccccccc}
\midrule
$N_c$,$N_d$,$N_v$   & \multicolumn{2}{c}{\textit{(30,3,2)}}    & \multicolumn{2}{c}{\textit{(30,3,3)}}     & \multicolumn{2}{c}{\textit{(30,3,5)}}    & \multicolumn{2}{c}{\textit{(50,3,3)}}  & \multicolumn{2}{c}{\textit{(50,3,5)}}& \multicolumn{2}{c}{\textit{(50,3,7)}}  \\
\cmidrule(lr){1-1}\cmidrule(lr){2-3}\cmidrule(lr){4-5}\cmidrule(lr){6-7}\cmidrule(lr){8-9}\cmidrule(lr){10-11}\cmidrule(lr){12-13}
       &  cost    & time     & cost     & time     & cost      & time      & cost      & time      & cost     & time     & cost    & time                                   \\
\cmidrule(lr){1-1}\cmidrule(lr){2-3}\cmidrule(lr){4-5}\cmidrule(lr){6-7}\cmidrule(lr){8-9}\cmidrule(lr){10-11}\cmidrule(lr){12-13}
Random & 2.43 & 0.42 & 1.84 & 0.61 & 1.48 & 0.94 & 2.18 & 1.43 & 1.62 & 2.39 & 1.47 & 2.62                     \\ 
Max-Min & 2.43 & 0.41 & 1.83 & 0.55 & 1.46 & 0.69 & 2.18 & 1.55 & 1.61 & 2.17 & 1.46 & 2.13                    \\

\bottomrule
\end{tabular}
}
\label{table:choosing 2 vehicle}
\end{table*}

\newpage
\subsection{Perturbation}
\label{appendix-subsub:perturbation}

NCE employs perturbation to increase performance. Perturbation is a commonly used strategy for enhancing the performance of meta-heuristics \cite{polat2015perturbation}. It is done by randomly \textit{perturbing} the solution and solving the problem with the perturbed solutions. This technique is beneficial to escape from the local optima. As described in \cref{alg:NCE-overall}, when falling into the local optima, NCE randomly selects two tours and performs a random exchange. We compared the performance of NCE according to perturbation. As shown in \cref{table:perturb}, the performance of NCE increases and converges as the number of perturbations $p$ increases. When $p=5$, the performance of NCE converges. Thus, we choose $p=5$ as the default hyperparameter of NCE.
\begin{table*}[h!]
\centering
\renewcommand{\arraystretch}{1.3} 

\caption{\textbf{Effect of perturbation}}

\vskip 0.15in
\resizebox{\textwidth}{!}{
\begin{tabular}{crcccccccccccccccc}
\midrule
$P$ & \multicolumn{2}{c}{\textit{0}}   & \multicolumn{2}{c}{\textit{1}}    & \multicolumn{2}{c}{\textit{2}}     & \multicolumn{2}{c}{\textit{3}}    & \multicolumn{2}{c}{\textit{5}}  & \multicolumn{2}{c}{\textit{7}}& \multicolumn{2}{c}{\textit{10}}& \multicolumn{2}{c}{\textit{20}} \\
\cmidrule(lr){1-1}\cmidrule(lr){2-3}\cmidrule(lr){4-5}\cmidrule(lr){6-7}\cmidrule(lr){8-9}\cmidrule(lr){10-11}\cmidrule(lr){12-13}\cmidrule(lr){14-15}\cmidrule(lr){16-17}
$N_c$,$N_d$,$N_v$         &  cost    & time     & cost     & time     & cost      & time      & cost      & time      & cost     & time     & cost    & time     & cost     & time     & cost    & time                               \\
\cmidrule(lr){1-1}\cmidrule(lr){2-3}\cmidrule(lr){4-5}\cmidrule(lr){6-7}\cmidrule(lr){8-9}\cmidrule(lr){10-11}\cmidrule(lr){12-13}\cmidrule(lr){14-15}\cmidrule(lr){16-17}
(30,3,2)       &2.50 &0.12  & 2.48 & 0.17 & 2.46 & 0.22 & 2.44 & 0.27 & 2.43 & 0.38& 2.43 & 0.49 & 2.42 & 0.69 & 2.41 & 1.34                     \\ 
(30,3,3)       & 1.89 & 0.16 & 1.86 & 0.22 & 1.84 & 0.30 & 1.84 & 0.35 & 1.83 & 0.55 & 1.82 & 0.61 & 1.81 & 0.81 & 1.81 & 1.42                     \\ 
(30,3,5)       & 1.49 &0.29  & 1.48 & 0.33 & 1.48 & 0.43 & 1.47 & 0.50 & 1.47 & 0.67 & 1.46 & 0.85 & 1.46 & 1.25 & 1.46 & 2.28                     \\ 

(50,3,3)    & 2.26 & 0.31 & 2.24 & 0.49 & 2.22 & 0.65 & 2.19 & 0.81 & 2.18 & 1.28 & 2.17 & 1.83 & 2.16 & 2.61 & 2.14 & 4.83                     \\ 
(50,3,5) & 1.66 & 0.52 & 1.64 & 0.77 & 1.63 & 0.94 & 1.62 & 1.24 & 1.61 & 1.97 & 1.61 & 2.59 & 1.60 & 3.61 & 1.59 & 6.53                     \\
(50,3,7)       & 1.48  &0.85  & 1.48 & 1.04 & 1.47 & 1.43 & 1.47 & 2.02 & 1.46 & 2.65 & 1.46 & 2.95 & 1.46 & 3.77 & 1.45 & 6.44                     \\

\bottomrule
\end{tabular}
}
\label{table:perturb}
\end{table*}

\newpage
\section{Training Detail}
\label{appendix:train-detail}

\paragraph{Dataset preparation} To train the cost-decrement prediction model $f_\theta(\cdot)$, we generate 50,000 random FMDVRP instances. The random instance is generated by first sampling the number of customer $N_c$ and depots $N_d$ from $\mathcal{U}(10,100)$ and $\mathcal{U}(2,9)$ and \textbf{$N_v=2$} respectively, and then sampling the 2D coordinates of the cities from $\mathcal{U}(0,1)$. As we set $N_v=2$, we generate two tours by applying the initial solution construction heuristics explained in \cref{subsec:NCE}. From $\tau_1, \tau_2$, we compute the true best cost-decrements of all feasible $(a_1, a_2)$ to prepare the training dataset. We generated 47,856,986 training samples from the 50,000 instances.

\paragraph{Hyperparameters} $f_\theta(\cdot)$ is parametrized via the GNN which employs five layers of the attentive embedding layer. We employ 4 layered MLPs to parameterize $\phi_e, \phi_w, \phi_n$ and $\phi_c$ whose hidden dimensions and activation units are 64 and Mish \cite{misra2019mish}. $f_\theta(\cdot)$ is trained to minimize Huber loss for three epochs via AdamW \cite{loshchilov2017decoupled} whose learning rate is fixed as $5\times 10^{-4}$.

\paragraph{Computing resources} We run all experiments on the server equipped with AMD Threadripper 2990WX CPU and Nvidia RTX 3090 GPU. We use a single CPU core for evaluating all algorithms.

\newpage

\section{Evaluation of the cost decrement model}
\label{appendix:train}

\begin{figure*}[h!]
 \centering
\includegraphics[width=0.9\textwidth]{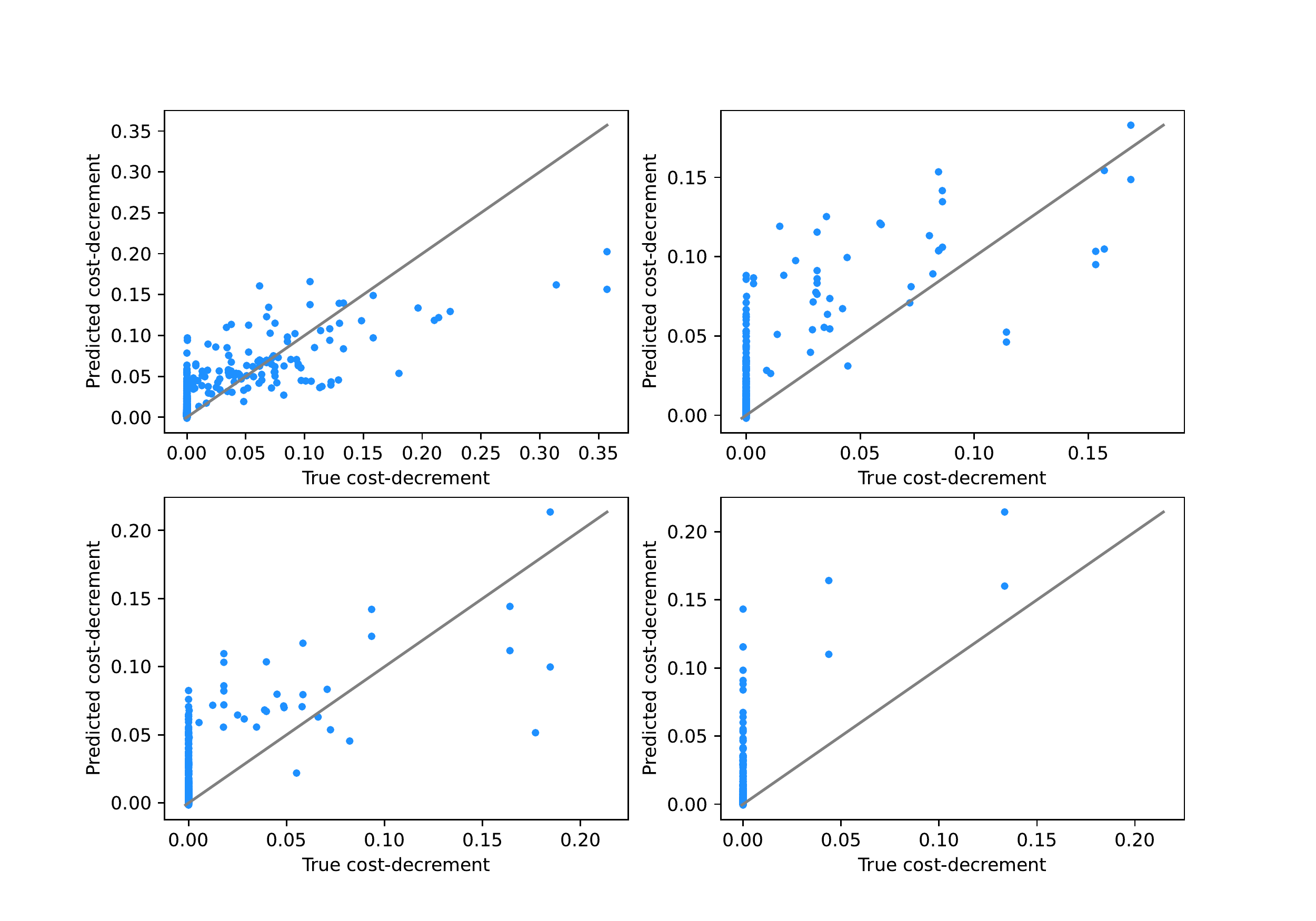}
 \caption{Predicted cost-decrements vs. true cost-decrements}
 \label{fig:precision}
 \end{figure*}

In this section, we evaluate the prediction accuracy of $f_{\theta}(\cdot)$. To evaluate $f_\theta(\cdot)$, we randomly generate 1,000 FMDVRP instances by sampling $N_C \sim \mathcal{U}(10, 100)$ and $N_D \sim \mathcal{U}(2, 9)$, and $(x,y) \sim \mathcal{U}(0,1)^2$. From the instances, we measure the ratio of existence of the argmax $(a_1,a_2)$ pair in the search candidate set whose size is $K$. As shown in \cref{table:precision}, when $K \geq 10$, NCE can find the argmax pair with at least $0.9$ probability. We also provide the 
results of the cost-decrement predictions and its corresponding cost. As shown in \cref{fig:precision}, $f_\theta(\cdot)$ well predicts the general tendency.

\begin{table*}[h]
\caption{\textbf{$f_{\theta}(\cdot)$ prediction performance test}}
\label{table:precision}
\centering
\footnotesize
\begin{tabular}{cccccc}
\midrule
$K$ & 1     & 3     & 5 & 10     & 20  \\
\cmidrule(lr){1-1}\cmidrule(lr){2-6}
argmax ratio (\%) & 42.9 & 71.3     & 78.6     & 90.9     & 97.4  \\\hline
\end{tabular}
\end{table*}

\newpage
\section{Comparison with full search}

\begin{figure}[h!]
\centering
\centerline{\includegraphics[width=\columnwidth]{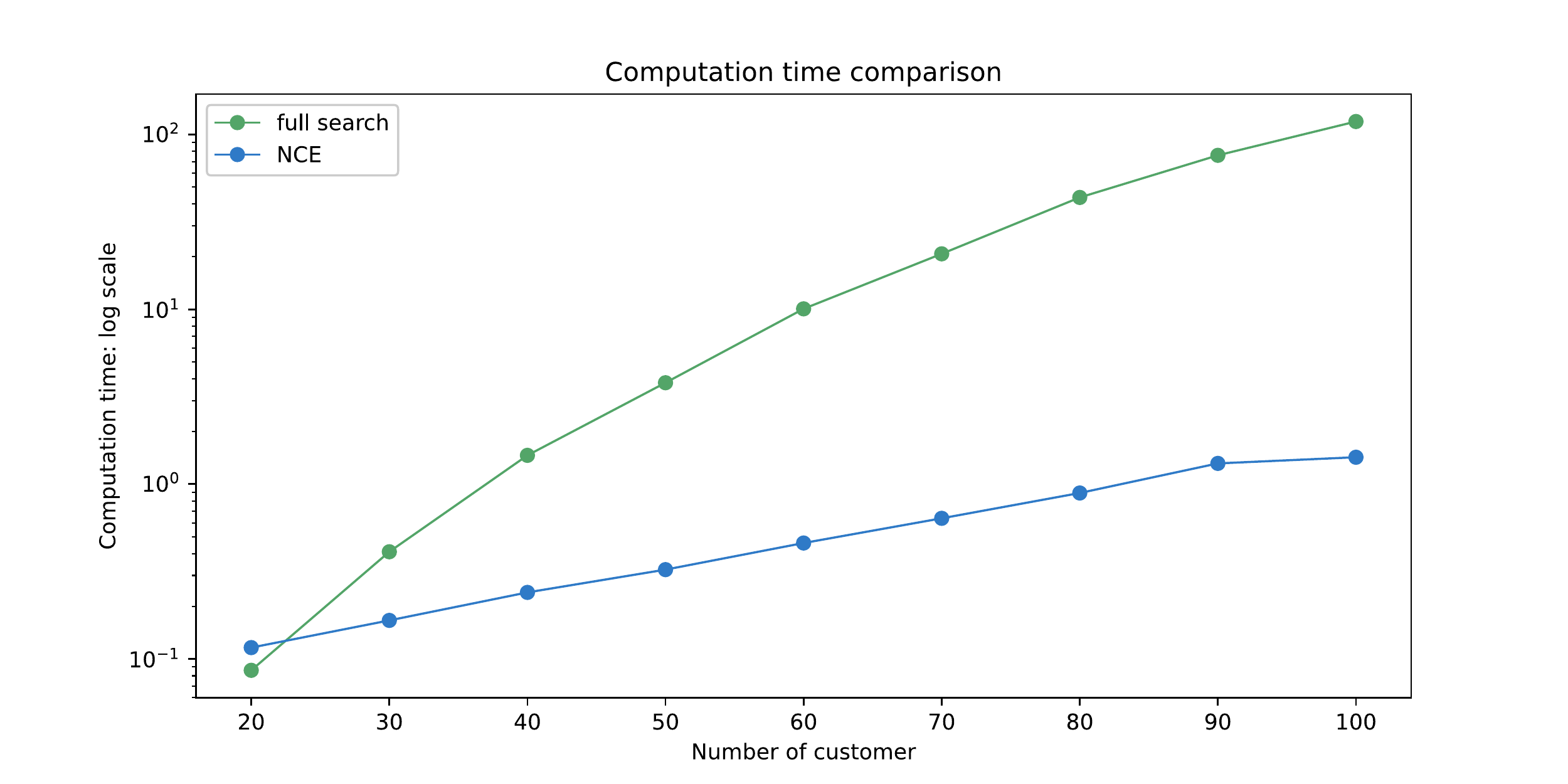}}
\vskip 0in
\caption{Computation speed comparison}
\label{fig:fullserach_comp_time}
\vskip 0in
\end{figure}

To verify NCE successfully amotrizes CE, we evaluate CE and NCE($K$=10, $p$=0) on FMDVRP. As the testing instances, we randomly generate 100 instances for each $N_c\in \{20,30,40,50,60,70,80,90,100\}$ with the fixed $N_d=3$ and $N_v=3$. As shown in \cref{table:full_search}, NCE shows nearly identical performances. On contrary, the computation speed of NCE is significantly faster than CE as shown in \cref{fig:fullserach_comp_time}.

\begin{table*}[h!]
\caption{\textbf{FMDVRP performance comparison of CE and NCE($K$=10, $p$=0)}}
\vskip 0.1in
\label{table:full_search}
\centering
\footnotesize
\begin{tabular}{cccccccccc}
\midrule
$N_d$ ,$N_v$ & \multicolumn{9}{c}{\textit{(3,3)}}  \\
\cmidrule(lr){1-1}\cmidrule(lr){2-10}
 $N_C$ &   \multicolumn{1}{c}{\textit{20}}    & \multicolumn{1}{c}{\textit{30}} & \multicolumn{1}{c}{\textit{40}}    & \multicolumn{1}{c}{\textit{50}}& \multicolumn{1}{c}{\textit{60}}    & \multicolumn{1}{c}{\textit{70}}& \multicolumn{1}{c}{\textit{80}} & \multicolumn{1}{c}{\textit{90}}& \multicolumn{1}{c}{\textit{100}}   \\
\cmidrule(lr){1-1}\cmidrule(lr){2-10}

CE  & 1.651    & 1.893     & 2.088    & 2.257     & 2.384 &2.531&2.695 &2.811&2.929\\
NCE & 1.651    & 1.891     & 2.088    & 2.262    & 2.390  &2.530&2.697 &2.806 &2.934\\
\bottomrule
\end{tabular}

\end{table*}

\vskip 0in

\section{Example solutions}	
\begin{figure}[!ht]
\begin{center}
\centerline{\includegraphics[width=\columnwidth]{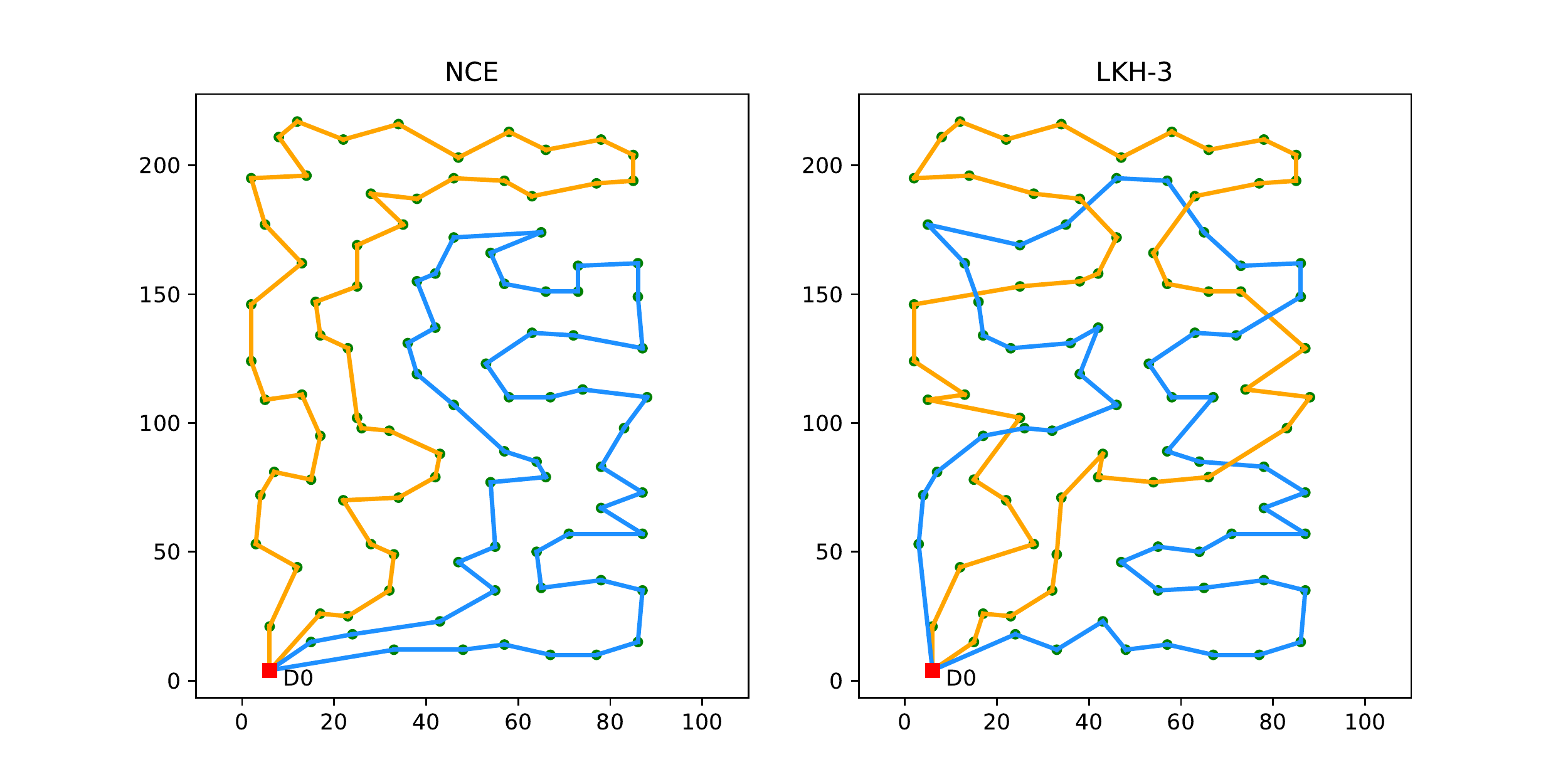}}
\vskip 0in
\caption{Rat99-2 solutions computed by NCE and LKH-3}
\label{fig:rat99}
\end{center}
\vskip 0in
\end{figure}

\begin{figure}[!ht]
\begin{center}
\centerline{\includegraphics[width=\columnwidth]{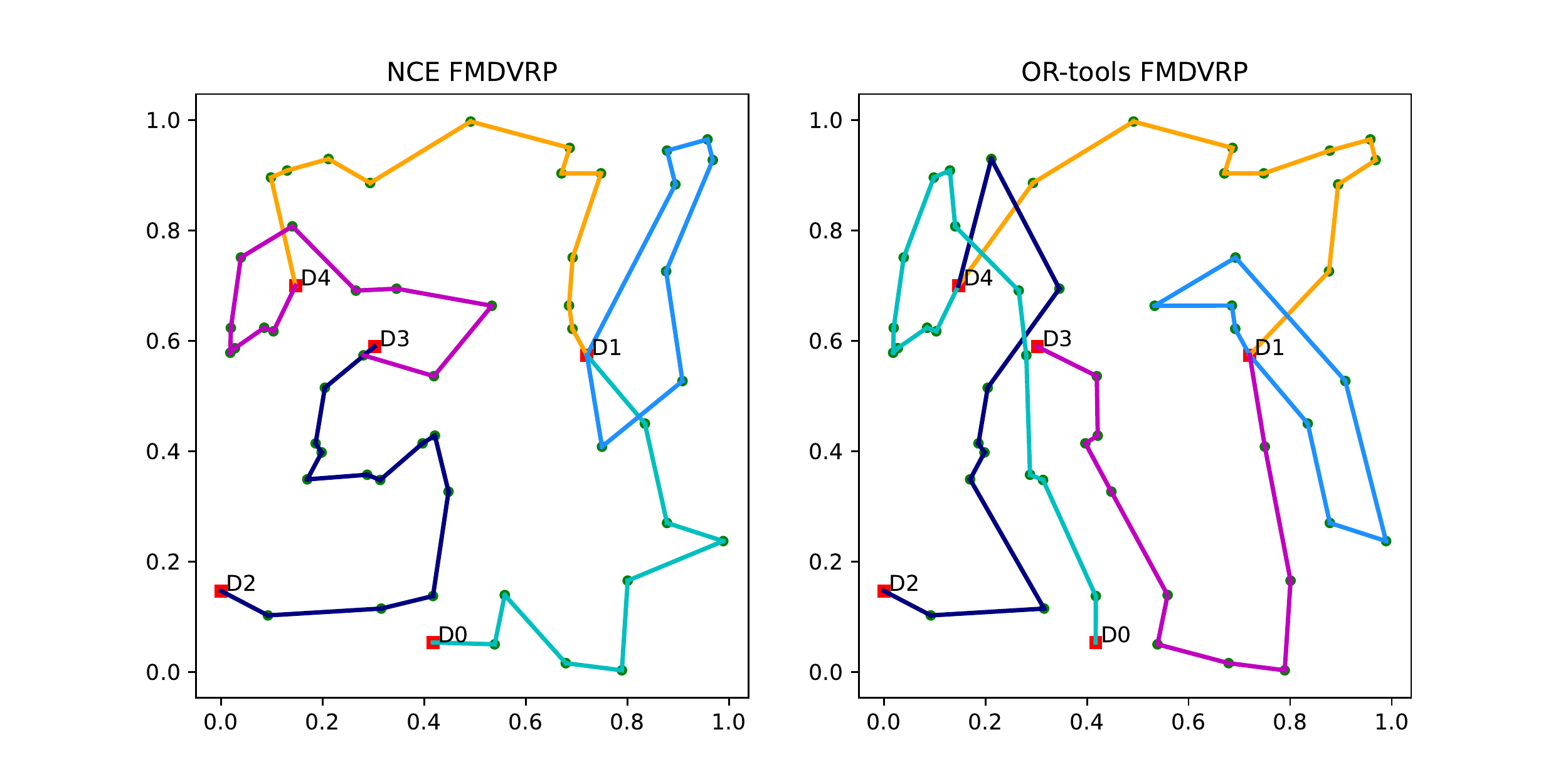}}
\vskip 0in
\caption{FMDVRP solutions computed by NCE and OR-tools}
\label{fig:fmd}
\end{center}
\vskip 0in
\end{figure}

\begin{figure}[!ht]
\begin{center}
\centerline{\includegraphics[width=\columnwidth]{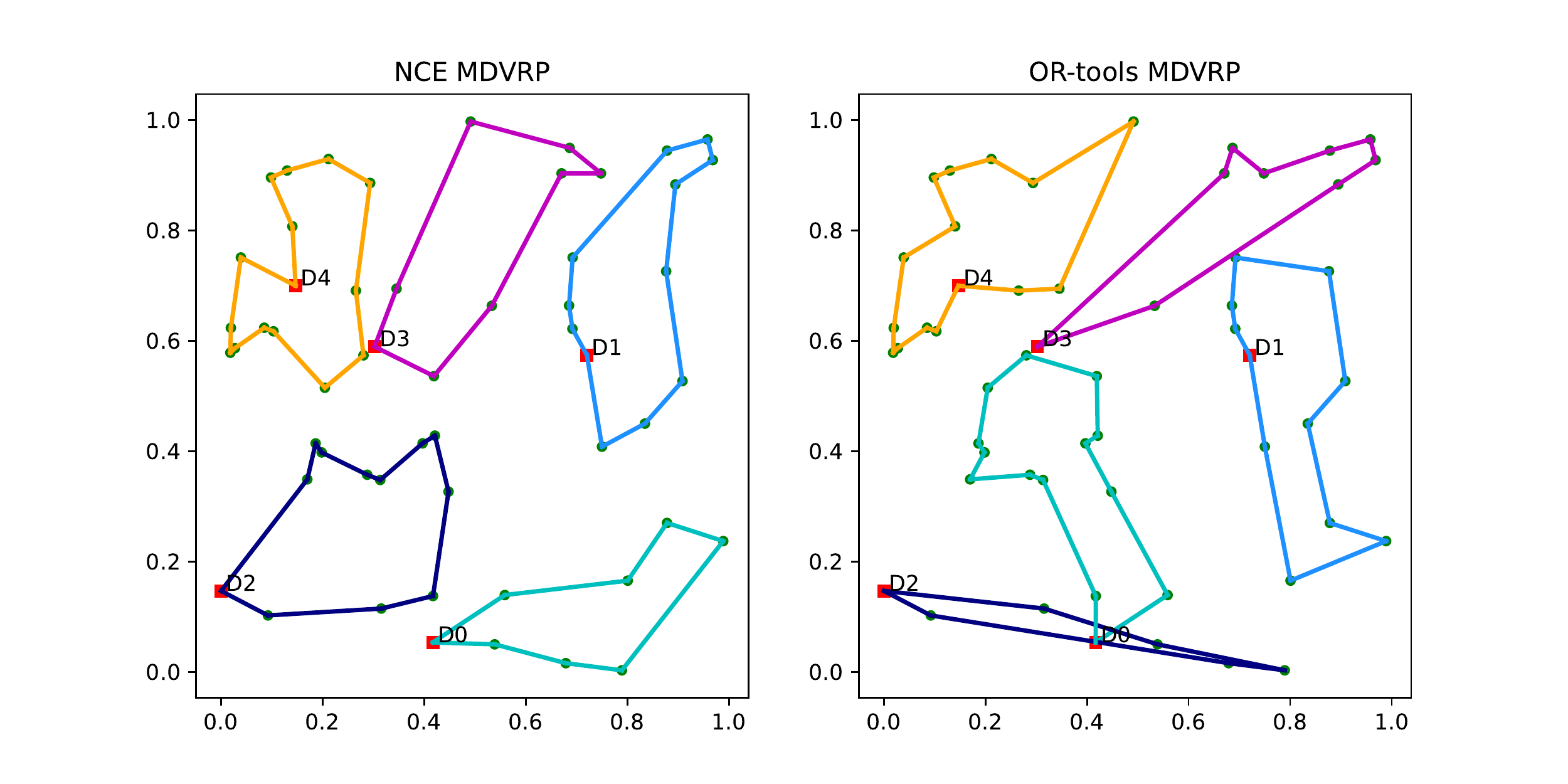}}
\vskip 0in
\caption{MDVRP solutions computed by NCE and OR-tools}
\label{fig:md}
\end{center}
\vskip 0in
\end{figure}

This section provides the routing examples. \cref{fig:rat99} shows the solution of Rat99-2 computed by LKH-3 and NCE. \cref{fig:fmd,fig:md} shows the solution of a FMDVRP and MDVRP instance computed by OR-Tools and NCE. 

\vskip 0in

\vskip 0in

\vskip 0in

\end{document}